\documentclass[lettersize,journal]{IEEEtran}
\usepackage{amsmath,amsfonts}
\usepackage{algorithmic}
\usepackage{algorithm}
\usepackage{array}
\usepackage[caption=false,font=normalsize,labelfont=sf,textfont=sf]{subfig}
\usepackage{textcomp}
\usepackage{stfloats}
\usepackage{url}
\usepackage{verbatim}
\usepackage{graphicx}
\usepackage{cite}
\usepackage{booktabs}
\usepackage{pifont}
\usepackage{multirow,hhline}
\usepackage[table]{xcolor}
\usepackage{soul}
\usepackage{arydshln}
\usepackage{bibunits}
\usepackage[colorlinks=true, 
            citecolor=blue,
            linkcolor=blue,
            urlcolor=blue]{hyperref}
\usepackage{color, xcolor}
\usepackage{gensymb}
\makeatletter
\newcommand{\Xhline}[1]{%
  \noalign{\ifnum0=`}\fi\hrule \@height #1 \futurelet\reserved@a\@xhline
}
\makeatother

\begin{document}

\title{Instance-Adaptive Keypoint Learning with Local-to-Global Geometric Aggregation for Category-Level Object Pose Estimation}

\author{Xiao~Zhang,
        Lu~Zou,
        Tao~Lu,
        Yuan~Yao,
	Zhangjin~Huang,~\IEEEmembership{Member,~IEEE,}
	and Guoping~Wang
	
\thanks{Xiao Zhang, Lu Zou, Tao Lu, and Yuan Yao are with Wuhan Institute of Technology, Wuhan 430205, China (e-mail: xiaozhang@stu.wit.edu.cn, lzou@wit.edu.cn, lut@wit.edu.cn, yaoyuan.usc@gmail.com). Zhangjin Huang is with University of Science and Technology of China, Hefei 230031, China (e-mail: zhuang@ustc.edu.cn). Guoping Wang is with Peking University, Beijing 100871, China (e-mail: wgp@pku.edu.cn). (\textit{Corresponding author: Lu Zou})}}

\maketitle

\begin{abstract}
Category-level object pose estimation aims to predict the 6D pose and size of previously unseen instances from predefined categories, requiring strong generalization across diverse object instances. Although many previous methods attempt to mitigate intra-class variations, they often struggle with instances (i.e., cameras) exhibiting complex geometries or significant deviations from canonical shapes. To address this issue, we propose INKL-Pose, a novel category-level object pose estimation framework that enables \underline{IN}stance-adaptive \underline{K}eyp\-oint \underline{L}earning with local-to-global geometric aggregation. Specifically, our method first predicts semantically consistent and geometrically informative keypoints using an Instance-Adaptive Keypoint Detector, then refines them: (1) a Local Keypoint Feature Aggregator capturing fine-grained geometries, and (2) a Global Keypoint Feature Aggregator using bidirectional Mamba for structural consistency. To enable bidirectional modeling in Mamba, we introduce a simple yet effective Feature Sequence Flipping strategy that preserves spatial coherence while constructing backward feature sequence. Additionally, we design a surface loss and a separation loss to encourage uniform coverage and spatial diversity in keypoint distribution. The resulting keypoints are mapped to a canonical space for 6D pose and size regression. Extensive experiments on CAMERA25, REAL275, and HouseCat6D show that INKL-Pose achieves state-of-the-art performance with 16.7M parameters and runs at 36 FPS on an NVIDIA RTX 4090D GPU.
\end{abstract}


\section{Introduction}
Object pose estimation is a fundamental yet challenging problem in computer vision and robotics, with broad applications in robotic grasping~\cite{wang2024oa, liu2024survey}, autonomous driving~\cite{qian20223d}, and 3D scene understanding~\cite{wu2024fairscene}. Existing approaches can be broadly classified into instance-level~\cite{di2021so, wang2019densefusion, tian2020robust} and category-level~\cite{wang2019normalized, lin2024instance, ren2025learning} methods. Instance-level methods rely on specific CAD models to estimate the poses of known objects. However, their heavy dependence on accurate CAD models makes them difficult to generalize to real-world scenarios, where objects are often novel and lack corresponding models. This severely limits their applicability to unseen instances within the same category or to objects with variations in shape and appearance. As a more flexible alternative, category-level object pose estimation aims to predict the pose and size of previously unseen instances from predefined object categories. Despite its potential, this task remains considerably more challenging due to the significant intra-class variations in object shape, texture, and structure.

\begin{figure}
        \centering
        \includegraphics[width=0.49\textwidth]{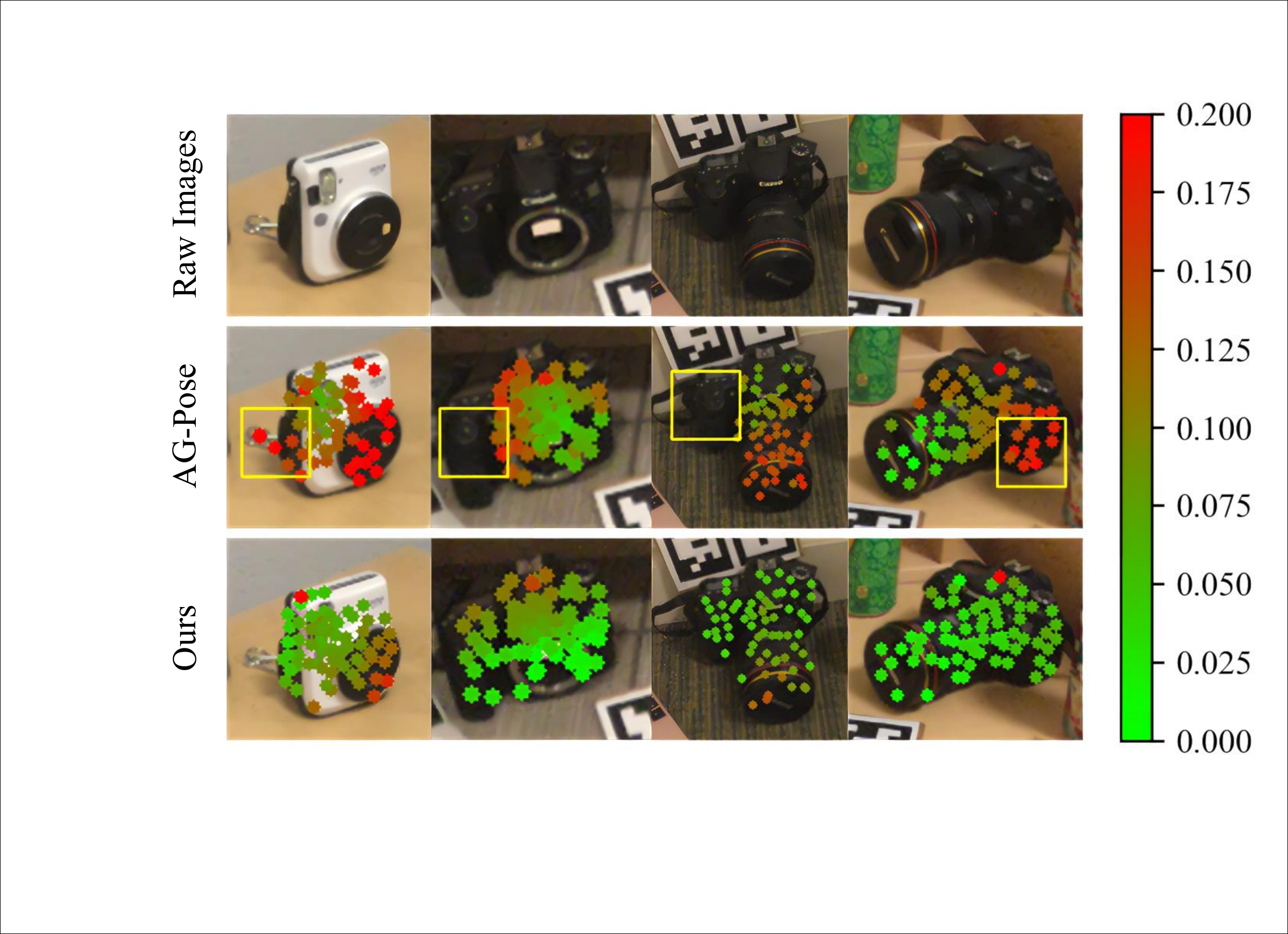}
        \caption{Visualization of keypoints in NOCS space predicted by AG-Pose~\cite{lin2024instance} and the proposed INKL-Pose on \textit{camera} instances with complex geometry. Keypoint colors indicate their NOCS errors, which are linearly mapped within the range \([0, 0.2]\) from green (error = 0) to red (error = 0.2). AG-Pose tends to produce clustered and outlier-prone keypoints with higher NOCS errors, whereas INKL-Pose generates spatially diverse and well-distributed keypoints that uniformly cover the object surface with significantly lower errors.}
        \label{fig:vis_kpt}
\end{figure}

\IEEEpubidadjcol

Recent advances in correspondence-based methods~\cite{tian2020shape, lin2022category, zou2024learning} have shown promising performance in category-level object pose estimation. However, these approaches typically rely on static priors that restrict their adaptability to novel object instances. To address this limitation, shape-prior-free methods~\cite{lin2023vi, chen2024secondpose, lin2024instance} have drawn increasing attention due to their enhanced generalization to novel object instances. Among these methods, AG-Pose~\cite{lin2024instance} proposes a keypoint-based framework that predicts representative keypoints for each instance and establishes keypoint-level correspondences, achieving significant improvements in handling structural discrepancies. Despite its success, AG-Pose derives keypoint features solely from local attention within fixed neighborhoods, combined with a simple global average pooling strategy. This design limits its capacity to capture fine-grained geometric cues and to model global structural consistency. Although AG-Pose incorporates losses to encourage uniformity and separation in keypoint distribution, their effectiveness diminishes when dealing with objects exhibiting complex or irregular geometries. As illustrated in Fig.~\ref{fig:vis_kpt}, the resulting keypoints often cluster in limited regions, leading to incomplete surface representation and substantial NOCS errors.

To address the above challenges, we propose INKL-Pose, a novel category-level object pose estimation method that enables instance-adaptive keypoint learning with local-to-global geometric aggregation. Specifically, we first extract and fuse RGB and point cloud features using a Pointwise Feature Encoder, and then generate a sparse set of semantically consistent and geometrically informative keypoints through an Instance-Adaptive Keypoint Detector. These keypoints are subsequently refined by two complementary modules: (1) a Local Keypoint Feature Aggregator that captures fine-grained geometric details from spatial neighborhoods, and (2) a Global Keypoint Feature Aggregator that utilizes a bidirectional Mamba architecture to model long-range dependencies while maintaining structural consistency across the object. To handle the unordered and non-causal nature of point clouds, we introduce a simple yet effective Feature Sequence Flipping strategy that facilitates bidirectional perception in the feature space while preserving the original spatial structure. Additionally, we design a surface loss and a separation loss to explicitly encourage uniform coverage and spatial diversity among keypoints. The refined keypoints are then mapped to a canonical space to regress the object’s 6D pose and size. As illustrated in Fig.~\ref{fig:vis_kpt}, even on complex \textit{camera} instances, the keypoints predicted by our INKL-Pose still achieve accurate and consistent surface coverage, and exhibit significantly lower NOCS errors compared to those predicted by AG-Pose~\cite{lin2024instance}. 

Overall, our contributions are summarized as follows:

\begin{itemize}
    \item We propose INKL-Pose, a novel keypoint-based framework for category-level 6D object pose estimation that enables instance-adaptive keypoint learning with local-to-global geometric aggregation.
    \item We introduce two complementary modules for keypoint feature refinement: a Local Keypoint Feature Aggregator that captures fine-grained geometric details from spatial neighborhoods, and a Global Keypoint Feature Aggregator that leverages bidirectional Mamba with a Feature Sequence Flipping strategy to enable coherent and globally consistent geometric perception.
    \item We design a surface loss and a separation loss to encourage keypoints to be uniformly and diversely distributed across the object surface, thereby enabling more effective capture of semantically consistent and geometrically informative features.
\end{itemize}

\section{Related Work}

In this section, we discuss the literature most relevant to our work from three perspectives: instance-level object pose estimation, category-level object pose estimation, and state space models.

\subsection{Instance-Level Object Pose Estimation}
Instance-level object pose estimation aims to infer the precise pose of an object based on its corresponding 3D CAD model. Existing approaches in this domain can be broadly divided into three categories: direct regression methods~\cite{di2021so, wang2019densefusion}, direct voting methods~\cite{tian2020robust, zhou2021pr} and dense 2D-3D correspondence methods~\cite{park2019pix2pose, xu2024rnnpose}.
Direct regression methods utilize end-to-end networks to directly predict the object's rotation and translation parameters from extracted visual features. Although this approach simplifies the pose estimation pipeline, it suffers from reduced accuracy in cluttered or occluded scenes due to the limitations of direct mapping.
Direct voting methods, on the other hand, estimate pose candidates and their corresponding confidence scores at the pixel or point level, ultimately selecting the pose with the highest confidence. While these methods are capable of capturing detailed local information and generally yield strong performance, the dense voting process can be computationally intensive and time-consuming.
Alternatively, the third category of instance-level methods adopts a correspondence-based paradigm, which first establishes dense 2D–3D correspondences between RGB images and the complete 3D CAD model, and subsequently estimates the object pose using algorithms such as Perspective-n-Point or least-squares optimization. Unlike direct regression or voting-based approaches, this method explicitly decouples feature matching from pose estimation, enabling more reliable, geometry-informed inference when accurate correspondences are available. However, their performance heavily relies on the availability of precise CAD models and robust feature descriptors, both of which are often difficult to obtain in real-world scenarios.

\subsection{Category-Level Object Pose Estimation}
Unlike instance-level approaches that rely on known CAD models, category-level 6D object pose estimation aims to estimate the pose and size of previously unseen instances from predefined object categories. This setting is inherently more general and challenging due to substantial intra-class variations among object instances. To address these challenges, NOCS~\cite{wang2019normalized} introduces the Normalized Object Coordinate Space, which maps different instances to a unified canonical frame, thereby mitigating variations in scale and geometry. The Umeyama algorithm~\cite{umeyama1991least} is then applied to recover the object pose via similarity transformation. To better handle intra-class variations, SPD~\cite{tian2020shape} proposes a deformation network that reconstructs object geometry by learning instance-specific deformations from category-level shape priors, followed by dense correspondence matching between the reconstructed canonical model and the observed point cloud. This paradigm has inspired a series of follow-up works~\cite{tian2020shape, lin2022category, zou2024learning}, which further explore the extraction of more robust features and the development of more efficient deformation strategies. However, the static nature of shape priors during both training and inference limits adaptability and hinders generalization to objects with substantial structural variations.

To overcome this limitation, Query6DoF~\cite{wang2023query6dof} introduces class-specific learnable queries that are jointly optimized with the pose estimation network, enabling the shape priors to adapt dynamically during training. In addition, shape-prior-free methods~\cite{liu2023net, lin2024clipose, chen2024secondpose, lin2024instance, lin2025cleanpose} have attracted increasing attention for their improved generalization to diverse and unseen object geometries within the same category. IST-Net~\cite{liu2023net} adopts an implicit spatial transformation framework that directly learns correspondences between camera-space and world-space features without relying on explicit priors. CLIPose~\cite{lin2024clipose} leverages a vision-language model to extract rich semantic features from both text and images, aligning them with point cloud representations via contrastive learning. SecondPose~\cite{chen2024secondpose} integrates geometric features with semantic embeddings from DINOv2~\cite{oquab2023dinov2}, supervised under SE(3)-consistent constraints to improve cross-instance generalization. SpherePose~\cite{ren2025learning} employs a shared spherical proxy representation to address semantic inconsistency arising from shape-dependent canonical spaces. CleanPose~\cite{lin2025cleanpose} integrates causal learning with knowledge distillation to address the spurious correlations introduced by confounding factors in pose estimation. More recently, AG-Pose~\cite{lin2024instance} proposes detecting instance-adaptive keypoints and establishing keypoint-level correspondences to enhance robustness against structural variations across object instances, achieving notable performance improvements.

Our work also focuses on establishing keypoint-level correspondences, but surpasses AG-Pose by introducing a novel keypoint detection method that incorporates explicit constraints on keypoint distribution. Furthermore, our method jointly captures fine-grained geometric details and global structural consistency, enabling the generation of semantically consistent and geometrically informative keypoints. This design provides strong geometric supervision and significantly enhances robustness to intra-class shape variations, ultimately leading to more accurate and generalizable pose estimation.

\subsection{State Space Models}
State Space Models (SSMs)~\cite{gu2022efficiently}, originating from the classical Kalman filter, have attracted increasing interest for their ability to model long-range dependencies with linear computational complexity, while also supporting parallel training. Variants such as S4 \cite{gu2022efficiently} introduce diagonal state matrices to reduce memory and computational overhead. Building upon this foundation, Mamba~\cite{gu2024mamba} further enhances the SSM framework by incorporating a selective mechanism that compresses context information based on input-dependent parameterization, improving both efficiency and performance. Since the original Mamba operates in a causal manner, where each output token can only attend to preceding tokens, subsequent work such as Vision Mamba~\cite{vim} introduces bidirectional Mamba to enable information flow in both directions and improve global interaction.

Motivated by these advances, recent studies~\cite{li2025enhancing, liang2024pointmamba, zhang2025point} have adapted the Mamba architecture for 3D point cloud analysis. For instance, PointMamba~\cite{liang2024pointmamba} transforms unordered point cloud data into ordered 1D sequences using Hilbert and Trans-Hilbert curves, preserving spatial continuity and semantic relevance. PCM~\cite{zhang2025point} further utilizes bidirectional Mamba with a point sequence flipping strategy to allow each point to access features from any other point in the cloud. Inspired by the success of Mamba in long-range sequence modeling, we explore its potential for category-level object pose estimation. In particular, we leverage bidirectional Mamba to model globally consistent geometric structures across varied object instances, and propose a novel bidirectional input sequence generation strategy tailored to the unordered and non-causal nature of point cloud data.

\section{Methodology}
\label{sec:method}
\subsection{Overview}
\label{sec:overview}

\begin{figure*}
    \centering
    \includegraphics[width=\textwidth]{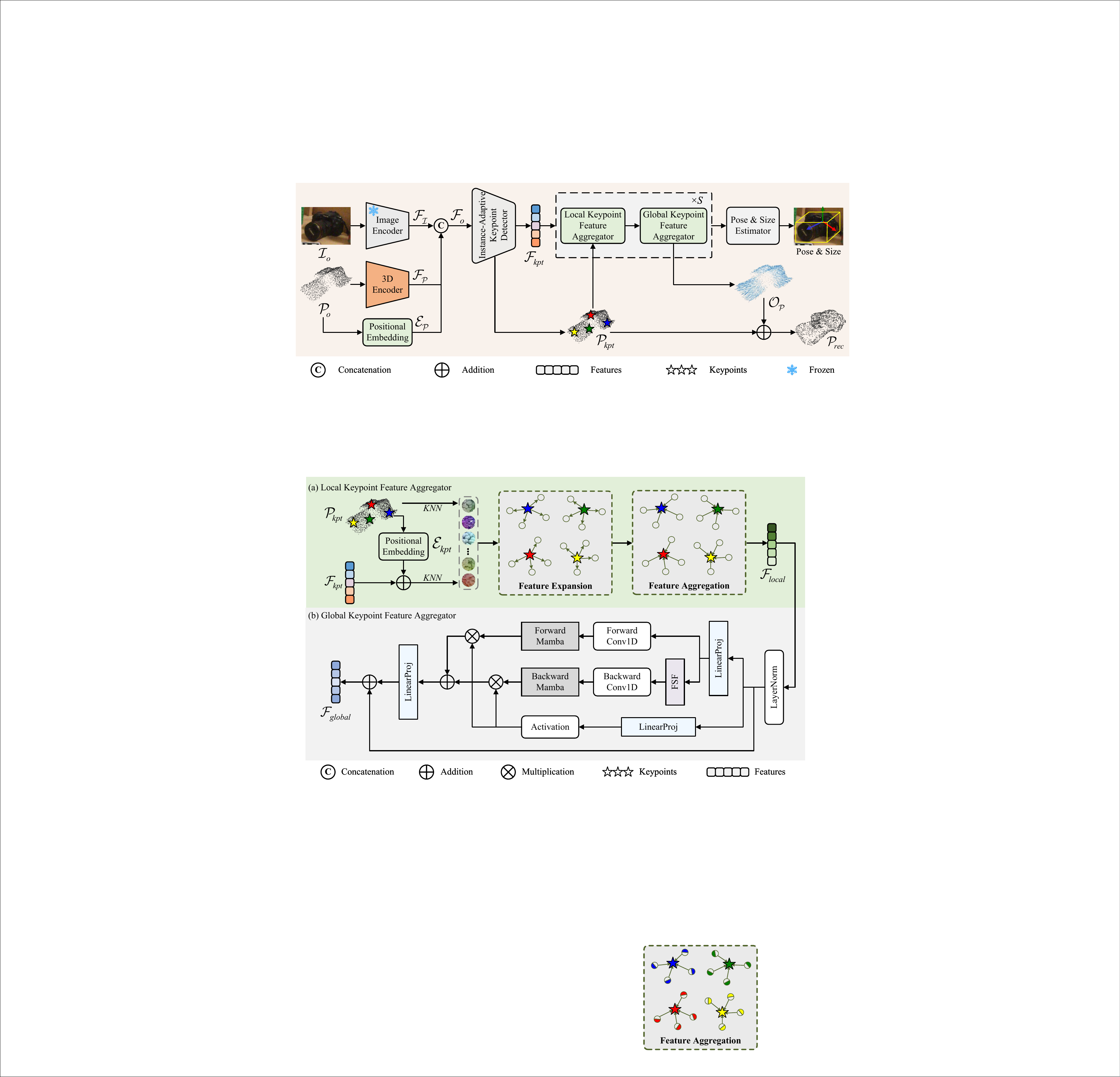}
    \caption{Overall framework of INKL-Pose. The Instance-Adaptive Keypoint Detector dynamically generates keypoints along with corresponding features using semantic and geometric information extracted by the Pointwise Feature Encoder. The Local Keypoint Feature Aggregator captures fine-grained local structures by aggregating neighborhood information from the point cloud. The Global Keypoint Feature Aggregator models long-range dependencies using bidirectional Mamba, extracting globally consistent geometric representations across instances. The Pose and Size Estimator predicts object rotation, translation, and size from the keypoints and their features.}
    \label{fig:framework}
\end{figure*}

In this section, we present the detailed architecture of the proposed INKL-Pose. As illustrated in Fig.~\ref{fig:framework}, our objective is to jointly estimate an object’s 3D rotation \( R \in \mathbb{R}^{3 \times 3} \), 3D translation \( t \in \mathbb{R}^3 \) and size \( s \in \mathbb{R}^3 \) from an input RGB image \( \mathcal{I}_o \) and its corresponding point cloud \( \mathcal{P}_o \).  
Specifically, INKL-Pose consists of five key stages. First, the Pointwise Feature Encoder (Section~\ref{sec:fea_extractor}) extracts instance features from both the RGB image and the point cloud, which are subsequently fused to facilitate cross-modal integration. Second, the fused features are processed by the Instance-Adaptive Keypoint Detector (Section~\ref{sec:kpt}) to predict a sparse set of keypoints and their associated features. Third, the Local Keypoint Feature Aggregator (Section~\ref{sec:LKFA}) enriches each keypoint with geometrically distinct local features. Fourth, the Global Keypoint Feature Aggregator (Section~\ref{sec:GKFA}) captures globally consistent geometric context across object instances. Finally, the Pose and Size Estimator (Section~\ref{sec:estimator}) takes the enhanced keypoints as input and directly regresses the object’s 6D pose and size.

\subsection{Pointwise Feature Encoder}
\label{sec:fea_extractor}

Given an RGB image \( \mathcal{I}_o \) and its corresponding point cloud \( \mathcal{P}_o \), the Pointwise Feature Encoder (PFE) extracts informative features from both modalities.  Specifically, semantic features \( \mathcal{F}_{\mathcal{I}} \in \mathbb{R}^{N \times d_1} \) are extracted from the RGB image using DINOv2~\cite{oquab2023dinov2}, while geometric features \( \mathcal{F}_{\mathcal{P}} \in \mathbb{R}^{N \times d_2} \) are obtained from the point cloud via PointNet++~\cite{qi2017pointnet++}. To emphasize relative geometric relationships and reduce the influence of absolute spatial positions, the point cloud is preprocessed by subtracting the mean of all coordinates. Although this operation benefits relative geometry modeling, it reduces the network’s sensitivity to absolute positional cues, which are essential for accurate pose estimation. To compensate for this loss, a Multi-Layer Perceptron (MLP) layer is employed to encode the original spatial coordinates into a latent positional embedding \( \mathcal{E}_{\mathcal{P}} \in \mathbb{R}^{N \times d_3} \). Finally, the semantic features \( \mathcal{F}_{\mathcal{I}} \), geometric features \( \mathcal{F}_{\mathcal{P}} \), and positional embedding \(\mathcal{E}_{\mathcal{P}}\) are concatenated and passed through an additional MLP layer for cross-modal feature fusion, yielding the final pointwise feature representation \( \mathcal{F}_o \in \mathbb{R}^{N \times d} \), which serves as input for subsequent network modules.

\subsection{Instance-Adaptive Keypoint Detector}
\label{sec:kpt}

Given the challenges of noise and incomplete observations in cluttered environments, we adopt the Instance-Adaptive Keypoint Detector (IAKD) from AG-Pose~\cite{lin2024instance} as our keypoint generation backbone. This detector formulates keypoint generation as a learnable, query-based process that dynamically generates keypoints from the most informative regions of each object. Rather than relying on fixed strategies like Farthest Point Sampling (FPS), which select keypoints solely based on spatial distances and fail to adapt to the geometric diversity across instances, the IAKD learns to predict instance-specific keypoints in a data-driven and context-aware manner. Specifically, we initialize a learnable keypoint query embedding \( \mathcal{Q}_{kpt} \) set to zero, providing a neutral starting point that avoids bias toward specific patterns. Through multi-head cross-attention and self-attention mechanisms, these query embeddings iteratively interact with the input features \( \mathcal{F}_{o} \) and themselves, progressively refining to capture instance-specific keypoint features. The keypoint probability distribution matrix \( \mathcal{W} \) is computed using cosine similarity followed by a softmax function:

\begin{equation}  
	\mathcal{W} = \text{Softmax}\left( \frac{\mathcal{Q}_{kpt} \cdot \mathcal{F}_o}{\|\mathcal{Q}_{kpt}\|_2 \cdot \|\mathcal{F}_{o}\|_2 + \varphi} \right),
\end{equation}

\noindent where \(\varphi\) is a small constant for numerical stability. Based on the matrix \( \mathcal{W} \), the keypoint coordinates $\mathcal{P}_{kpt} \in \mathbb{R}^{N_{kpt} \times 3}$ and keypoint features $\mathcal{F}_{kpt} \in \mathbb{R}^{N_{kpt} \times d} $ are computed as weighted averages of input coordinates \( \mathcal{P}_o \) and fused features \( \mathcal{F}_{o} \):

\begin{equation}  
	\mathcal{P}_{kpt} = \mathcal{W} \mathcal{P}_o, \quad \mathcal{F}_{kpt} = \mathcal{W} \mathcal{F}_o.
\end{equation}

To ensure the predicted keypoints are both well-distributed across the object surface and geometrically diverse, we introduce two complementary loss functions. First, the surface loss \( \mathcal{L}_{surf} \) encourages uniform surface coverage by minimizing the Chamfer Distance between the predicted keypoints \( \mathcal{P}_{kpt} \) and a set of reference points \( \mathcal{P}_{fps} \), which are uniformly sampled from the input point cloud using FPS:

\begin{equation}
	\begin{aligned}
		\mathcal{L}_{surf}(\mathcal{P}_{fps}, \mathcal{P}_{kpt})
		=& \, \frac{1}{{N_{fps}}} \sum_{x_i \in \mathcal{P}_{fps}} \min_{y_i \in \mathcal{P}_{kpt}} \|x_i - y_i\|_2^2+ \\
		& \, \frac{1}{{N_{kpt}}} \sum_{y_i \in \mathcal{P}_{kpt}} \min_{x_i \in \mathcal{P}_{fps}} \|y_i - x _i\|_2^2
	\end{aligned}.
\end{equation}

Second, the separation loss \(\mathcal{L}_{sep}\) promotes geometric diversity by encouraging each keypoint to be spatially distinct from its local neighbors. Specifically, it maximizes the average pairwise distance between each predicted keypoint and its \(M\)-nearest neighbors in Euclidean space:

\begin{equation} \mathcal{L}_{sep} = \frac{1}{\max\left( \frac{1}{N_{kpt}} \sum_{i=1}^{N_{kpt}} \frac{1}{M} \sum_{j=1}^{M} \| \mathcal{P}_{kpt}^i - \text{NN}_j(\mathcal{P}_{kpt}^i) \|_2, \epsilon \right)}, \end{equation}

\noindent where \(\text{NN}_j(\mathcal{P}_{kpt}^i)\) denotes the \(j\)th nearest neighbor of the \(i\)th keypoint, and \(M\) is the number of neighbors considered. In our experiments, we set \(M = 2\), which provides a lightweight yet effective constraint to mitigate keypoint clustering while maintaining flexibility in spatial layout. The denominator is lower-bounded by a constant \(\epsilon\) to ensure numerical stability.

By jointly optimizing these two loss terms, the predicted keypoints are encouraged to be uniformly distributed and spatially diverse, thus establishing a robust foundation for accurate pose estimation. The effectiveness of these loss functions is demonstrated through ablation studies in Section~\ref{sec:ablation}.

\begin{figure*}[!t]
	\centering
	\includegraphics[width=\textwidth]{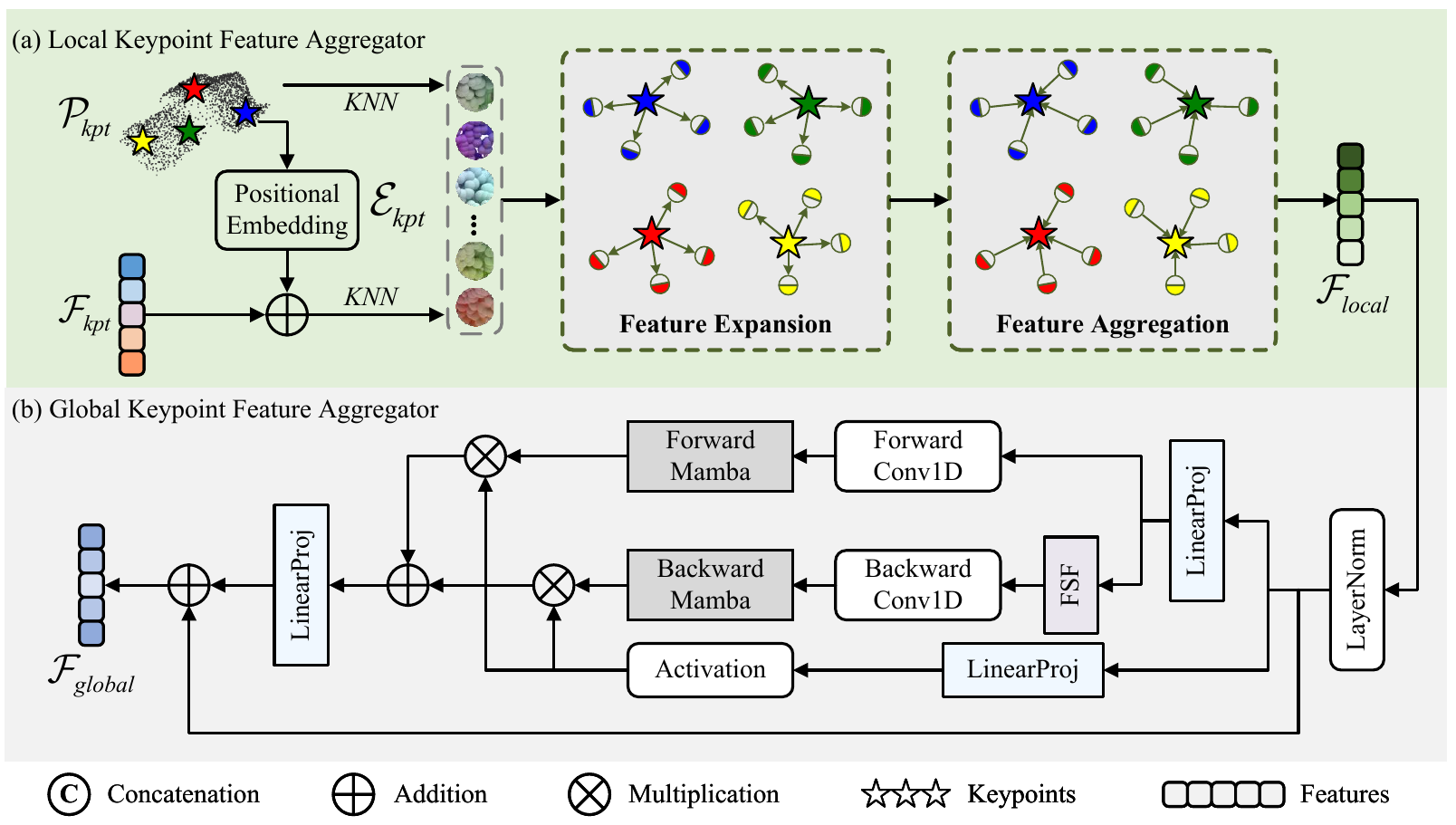}
	\caption{Illustration of the Local Keypoint Feature Aggregator (LKFA) and Global Keypoint Feature Aggregator (GKFA). (a) LKFA leverages KNN-based neighborhood construction and positional encoding, followed by feature expansion and aggregation, to capture fine-grained geometric details from local neighborhoods. (b) GKFA applies feature sequence flipping to obtain a backward sequence and leverages bidirectional Mamba to model long-range dependencies and global shape structures by integrating both original and flipped feature sequences.}
	\label{fig:keypoint_agg}
\end{figure*}

\subsection{Local Keypoint Feature Aggregator}
\label{sec:LKFA}

While the extracted keypoints effectively localize salient regions of an object, their sparse nature may result in the loss of fine-grained geometric details. To address this limitation, we propose a Local Keypoint Feature Aggregator (LKFA) that enriches each keypoint by integrating geometrically distinct local features from its surrounding neighborhood. 

As illustrated in Fig.~\ref{fig:keypoint_agg}(a), we first utilize a coordinate encoding function, implemented as an MLP layer, to project the keypoint coordinates into a high-dimensional feature space, producing positional embeddings \(\mathcal{E}_{kpt} \in \mathbb{R}^{N_{kpt} \times d}\). These embeddings are then added to the initial keypoint features \(\mathcal{F}_{kpt}\) to enhance spatial awareness:

\begin{equation}
	\mathcal{E}_{kpt} = \text{MLP}(\mathcal{P}_{kpt}), \quad \mathcal{F}_{kpt} = \mathcal{F}_{kpt} + \mathcal{E}_{kpt}.
\end{equation}

Next, for each keypoint, we construct a local neighborhood in both coordinate and feature spaces using the \( K \)-nearest neighbor ($K\!N\!N$) algorithm. This yields coordinate neighborhoods \(\mathcal{P}_K \in \mathbb{R}^{N_{kpt} \times K \times 3}\) and feature neighborhoods \(\mathcal{F}_K \in \mathbb{R}^{N_{kpt} \times K \times d}\), where \( K \) is the neighborhood size. We then concatenate the spatial and feature descriptors and apply standard normalization to obtain fused local group features \(\mathcal{F}_K' \in \mathbb{R}^{N_{kpt} \times K \times (d+3)}\), which jointly encode geometric structure and semantic information:

\begin{equation}
	\begin{aligned}
		\mathcal{F}_K' = \text{Norm}(\text{Concat}(\mathcal{F}_K, \mathcal{P}_K)).
	\end{aligned}
\end{equation}

To further enhance the representational capacity of each keypoint, we concatenate the normalized neighbor features \(\mathcal{F}'_{K}\) with the corresponding central keypoint feature \(\mathcal{F}_{kpt}\), resulting in expanded representations \(\mathcal{F}''_{K} \in \mathbb{R}^{N_{kpt} \times K \times (2\times d+3)} \). This operation allows each keypoint to incorporate both its own feature and contextual information from its spatial neighborhood. We then apply an affine transformation to \(\mathcal{F}''_{K}\), parameterized by learnable weights \(\alpha\) and \(\beta\), to capture potential rigid transformations in local geometry and adaptively refine the feature distribution. The transformed feature set \(\hat{\mathcal{F}}_K \in \mathbb{R}^{N_{kpt} \times K \times (2\times d+3)} \) is computed as follows:

\begin{equation}
	\begin{aligned}
		\mathcal{F}_K'' = \text{Concat}(\mathcal{F}_K', \mathcal{F}_{kpt}), \quad \hat{\mathcal{F}}_K = \mathbf{\alpha} \odot \mathcal{F}_K'' + \mathbf{\beta},
	\end{aligned}
\end{equation}

\noindent  where \( \odot \) denotes element-wise multiplication. 

A max-pooling operation is then applied across each local neighborhood to aggregate the most discriminative geometric and semantic cues into the central keypoint representation. Finally, an MLP layer is used to refine the output and produce the final local feature representation \( \mathcal{F}_{local} \in \mathbb{R}^{N_{kpt} \times d} \):

\begin{equation}
	\begin{aligned}
		\mathcal{F}_{local} = \text{MLP}(\text{MaxPool}(\mathbf{\hat{\mathcal{F}}}_K)).
	\end{aligned}
\end{equation}

By aggregating information from spatially coherent local regions, our LKFA effectively captures geometrically distinct local variations, enhancing the descriptive capacity of each keypoint and laying a solid foundation for robust global perception.

\subsection{Global Keypoint Feature Aggregator}
\label{sec:GKFA}

While the Local Keypoint Feature Aggregator (LKFA) effectively captures fine-grained geometric details, it lacks the capacity to model global structural context and long-range dependencies among keypoints. To address this limitation, we introduce a Global Keypoint Feature Aggregator (GKFA), which enhances semantic consistency by leveraging SSMs.

As illustrated in Fig.~\ref{fig:keypoint_agg}(b), GKFA employs a bidirectional Mamba-based architecture~\cite{gu2024mamba} to efficiently model long-range dependencies and capture object-level structural information. Compared to traditional Transformers, which suffer from quadratic complexity with respect to sequence length, Mamba operates in linear time, making it a more scalable and efficient choice for processing keypoint sequences in point clouds.

To fully exploit Mamba's bidirectional modeling capabilities, it is necessary to construct both forward and backward feature sequences. A common practice in prior works is to reverse the input sequence order, a technique we refer to as Point Sequence Flipping (PSF). However, point cloud data lack an inherent spatial order, and reversing the index sequence may place geometrically unrelated keypoints next to each other, thereby disrupting the continuity of feature learning. To address this issue, we propose a simple yet effective strategy termed Feature Sequence Flipping (FSF). Rather than reordering keypoints, FSF reverses the feature channels within each keypoint vector while preserving the original spatial arrangement of the point cloud. Formally, given a keypoint feature vector \(\mathcal{F}_{local}(i) \in \mathbb{R}^d\) expressed as \([f_1(i), f_2(i), \ldots, f_d(i)]\), we define FSF as:

\begin{equation}
    \text{FSF}\left(\mathcal{F}_{\text{local}}\left(i\right)\right) = \left[ f_d(i), f_{d-1}(i), \ldots, f_1(i) \right].
\end{equation}

Since different feature channels encode complementary semantics such as geometry or part cues, this reversal introduces a novel perspective on feature dependencies while maintaining the geometric integrity of the point cloud.

In the GKFA, both the original (forward) and FSF-processed (backward) feature sequences are fed into two parallel Mamba blocks. Each block independently captures long-range dependencies within its respective sequence. The outputs from the two Mamba blocks are then aggregated via element-wise summation, forming the final global keypoint features:
\begin{equation}
    \mathcal{F}_{global}=\text{SSM}\left(\mathcal{F}_{local}, \text{FSF}\left(\mathcal{F}_{local}\right)\right) + \mathcal{F}_{local}.
\end{equation}

This design allows the model to incorporate both the original and flipped feature perspectives, promoting more comprehensive global context learning while maintaining computational efficiency and stability.

To ensure that the aggregated keypoints retain meaningful and expressive geometric structure, we utilize the reconstruction-based supervision mechanism~\cite{lin2024instance}. Specifically, we employ an MLP layer to regress dense positional offsets from the global keypoint features. Given \(\mathcal{F}_{global} \in \mathbb{R}^{N_{kpt} \times d}\), we repeat each feature vector \( m = \lfloor N_{rec} / N_{kpt} \rfloor \) times to construct an expanded representation \(\mathcal{F}_{global}^\prime \in \mathbb{R}^{N_{rec} \times d}\), where \(N_{rec}\) is the predefined number of reconstruction points (set to 960 in our experiments). The MLP predicts a set of 3D offsets \(\mathcal{O}_\mathcal{P} \in \mathbb{R}^{N_{rec} \times 3}\):

\begin{equation}
    \mathcal{O}_\mathcal{P} = \text{MLP}(\mathcal{F}_{global}^\prime).
\end{equation}

Similarly, the keypoint coordinates \(\mathcal{P}_{kpt} \in \mathbb{R}^{N_{kpt} \times 3}\) are repeated \(m\) times to obtain \(\mathcal{P}_{kpt}^\prime \in \mathbb{R}^{N_{rec} \times 3}\). The reconstructed point cloud \(\mathcal{P}_{rec}\) is then computed as: \(\mathcal{P}_{rec} = \mathcal{P}_{kpt}^\prime + \mathcal{O}_\mathcal{P}.\)

To ensure geometric consistency between the reconstructed and observed point clouds, we adopt a Chamfer Distance-based similarity loss \(\mathcal{L}_{sim}\), which measures the bidirectional alignment between \(\mathcal{P}_{rec}\) and \(\mathcal{P}_o\):

\begin{equation}
	\begin{aligned}
		\mathcal{L}_{sim}(\mathcal{P}_o, \mathcal{P}_{rec})
		=& \, \frac{1}{N_{pts}} \sum_{x_i \in \mathcal{P}_o} \min_{y_j \in \mathcal{P}_{rec}} \|x_i - y_j\|_2^2+ \\
		& \frac{1}{N_{rec}} \sum_{y_j \in \mathcal{P}_{rec}} \min_{x_i \in \mathcal{P}_o} \|y_j - x_i\|_2^2.
	\end{aligned}
\end{equation}

To further improve representational capacity, we stack multiple LKFA and GKFA modules, refining keypoint features hierarchically. Our ablation studies (Section~\ref{sec:ablation}) show that stacking $S=12$ blocks yields the best overall performance.

\subsection{Pose and Size Estimator} 
\label{sec:estimator}

After obtaining semantically consistent and geometrically informative keypoints and their associated features, we adopt a standard Pose and Size Estimator, as in previous work~\cite{lin2022category,lin2024instance}, to infer the object's 6D pose and size. Specifically, we first transform the predicted keypoint coordinates into the NOCS space using the ground-truth rotation $\textit{R}_{gt}$, translation $\textit{t}_{gt}$, and scale $\textit{s}_{gt}$:
\begin{equation}
    \mathcal{P}_{kpt}^{gt}=\left\{\frac{1}{\|\textit{s}_{gt}\|_2}(x_i-\textit{t}_{gt})\textit{R}_{gt}\mid x_i\in \mathcal{P}_{kpt}\right\}.
\end{equation}

We then predict the NOCS keypoints from the extracted features:
\begin{equation}
    \mathcal{P}_{kpt}^{nocs}=\text{MLP}(\text{Self-Attn}(\mathcal{F}_{kpt})).
\end{equation}

To supervise this mapping, we adopt a smooth $\mathcal{L}_{1}$ loss:
\begin{equation}
    \mathcal{L}_{map} = \frac{1}{N_{kpt}} \sum_{i=1}^{N_{kpt}} \text{Smooth-}L_1\left(\mathcal{P}_{kpt}^{nocs}(i) - \mathcal{P}_{kpt}^{gt}(i)\right),
\end{equation}

\begin{equation}
    \text{Smooth-}L_1(x) = 
    \begin{cases}
        0.5 x^2, & \text{if } |x| < 1 \\
        |x| - 0.5, & \text{otherwise}.
    \end{cases}
\end{equation}

Finally, we regress the object rotation $R_{pre}$, translation $t_{pre}$, and size $s_{pre}$ using parallel MLPs based on the concatenated keypoint information: $\mathcal{F}_{pose}=\text{Concat}(\mathcal{P}_{kpt}, \mathcal{P}_{kpt}^{nocs}, \mathcal{F}_{kpt})$. We apply an $\mathcal{L}_{2}$ loss to supervise these predictions:
\begin{equation}
    \mathcal{L}_{pose} =\|R_{pre} - R_{gt} \|_2 + \|t_{pre} - t_{gt} \|_2 + \| s_{pre} - s_{gt} \|_2.
\end{equation}

\subsection{Overall Loss Function}

In summary, the overall loss function is as below:
\begin{equation}
	\begin{aligned}
		\mathcal{L}_{all}=\, &\omega_{sep}\mathcal{L}_{sep}+\omega_{surf}\mathcal{L}_{surf}+ \\ &\omega_{sim}\mathcal{L}_{sim}+\omega_{map}\mathcal{L}_{map}+\omega_{pose}\mathcal{L}_{pose},
	\end{aligned}
\end{equation}
where $\omega_{sep}$, $\omega_{surf}$, $\omega_{sim}$, $\omega_{map}$, and $\omega_{pose}$ are hyperparameters used to balance the contributions of separation loss, surface loss, similarity loss, mapping loss and pose loss.

\section{Experiments}
\label{sec:exp}
\subsection{Experimental Setup}
\subsubsection{Datasets}

Following previous works~\cite{lin2024instance,ren2025learning}, we evaluate the effectiveness of the proposed INKL-Pose on three widely used benchmarks: CAMERA25~\cite{wang2019normalized}, REAL27\-5~\cite{wang2019normalized}, and the more challenging HouseCat6D~\cite{jung2024housecat6d} dataset. CAMERA25 is a large-scale synthetic dataset covering six object categories, containing 300k synthetic RGB-D images with objects rendered on virtual backgrounds, of which 25k images are used for evaluation. REAL275 serves as a real-world counterpart to CAMERA25, covering the same six categories. The training set comprises 4.3k images captured from 7 scenes, while the test set includes 2.75k images from 6 different scenes, with three distinct object instances per category. HouseCat6D is a more complex, real-world multimodal dataset featuring 10 object categories. The training set contains 20k RGB-D images collected across 34 indoor scenes, and the test set includes 3k images from 5 additional scenes. Notably, this dataset poses greater challenges due to its marker-free capture process, dense clutter, and severe occlusion.

\subsubsection{Evaluation Metrics}

Following previous works~\cite{lin2024instance,ren2025learning}, we evaluate the performance of INKL-Pose using two widely adopted metrics.

\begin{itemize}
    \item IoU\textsubscript{X}. This metric calculates the 3D Intersection-over-Union (IoU) between the predicted and ground-truth bounding boxes. A prediction is considered correct if the IoU exceeds a specified threshold. Following standard protocol, we report IoU\textsubscript{50} and IoU\textsubscript{75} on CAMERA25 and REAL275, and IoU\textsubscript{25} and IoU\textsubscript{50} on the more challenging HouseCat6D dataset.
    \item $n$\textdegree$m$cm. This metric evaluates the 6D pose accuracy by jointly considering rotation and translation errors. A prediction is regarded as correct if the rotation error is less than $n$\textdegree\ and the translation error is below $m$ cm. Following standard protocol, we report results under four commonly used thresholds: 5\textdegree2cm, 5\textdegree5cm, 10\textdegree2cm, and 10\textdegree5cm across all three datasets.
\end{itemize}

\subsubsection{Implementation Details}

We utilize the off-the-shelf Mask R-CNN~\cite{he2017mask} to generate instance masks and resize all input images to 224 \(\times\) 224 pixels. For each corresponding object instance, we uniformly sample 1,024 points from the segmented point cloud. To enhance robustness, we apply data augmentation strategies consistent with prior works~\cite{lin2024instance, lin2022category}, including random rotations (within 0 – 20 degrees), translations (in the range of –0.02 to 0.02 meters), and scalings (from 0.8 to 1.2). For network configuration, the feature dimensions are set as follows: \( d_1\) = 128, \( d_2\) = 128 , \( d_3\) = 64 , and the fused feature dimension is \( d = 256 \). In the local feature aggregation stage, the neighborhood size \(K\) is set to 4. Additionally, the entire feature aggregation module consists of \(S\) = 12 stacked stages, each comprising both local and global aggregation components to progressively refine the keypoint features. In the keypoint generation stage, the network predicts 96 keypoints per object instance. For keypoint supervision, we sample \(N_{fps}\) = 120 reference points to ensure sufficient surface coverage and spatial alignment with the predicted keypoints. In the separation loss, the number of nearest neighbors is set to \(M\) = 2, which provides a lightweight yet effective constraint that mitigates keypoint clustering while preserving flexibility in spatial distribution. In the dense point cloud reconstruction stage, each keypoint is repeated 10 times, yielding \(N_{rec}\) = 960 reconstructed points. This setting closely matches the resolution of the input point cloud (1,024 points), providing a stable foundation for reconstruction supervision without incurring excessive computational cost. 

Regarding optimization, the weights of the loss terms are set as \( \{ w_{sep}, w_{surf}, w_{sim}, w_{map}, w_{pose} \}
= \{ 10.0,\allowbreak 10.0,\allowbreak 15.0,\allowbreak 2.0,\allowbreak 0.3 \} \). The model is trained using the Adam optimizer with a cyclical learning rate schedule~\cite{smith2017cyclical}, where the learning rate varies between \( 2 \times 10^{-5} \) and \( 5 \times 10^{-4} \). All experiments are conducted on a single NVIDIA RTX 4090D GPU with a batch size of 64. For CAMERA25 and REAL275, we train the model using both synthetic and real-world data with a synthetic-to-real ratio of 3:1. For HouseCat6D, the model is trained exclusively on its corresponding dataset.

\renewcommand{\arraystretch}{1.15}
\begin{table*}
    \centering
    \belowrulesep=0pt
    \aboverulesep=0pt
    \setlength{\abovecaptionskip}{0cm}
     \caption{Quantitative comparisons with state-of-the-art methods on the REAL275 dataset~\cite{wang2019normalized}. The best results are in \textbf{bold}.  The symbol $^\dagger$ indicates that only the trainable parameter count is reported. The symbol $-$ indicates that the data is not available or not reported.}
	\resizebox{0.9\textwidth}{!}{\begin{tabular}{c|c|c c|c c c c c} 
    \toprule
			\multirow{2}{*}{Method} & \multirow{2}{*}{Prior} & \multicolumn{6}{c}{REAL275 (\(\uparrow\))} & \#Params \\
			\cline{3-8} 
			& & $\mathrm{IoU_{50}}$ & $\mathrm{IoU_{75}}$ & $5^{\circ}2\mathrm{cm}$ & $5^{\circ}5\mathrm{cm}$ & $10^{\circ}2\mathrm{cm}$ & $10^{\circ}5\mathrm{cm}$ &(M) \(\downarrow\)\\
			\midrule
			SPD [ECCV'20]~\cite{tian2020shape}       & \ding{51} & 77.3 & 53.2 & 19.3 & 21.4 & 43.2 & 54.1 &18.3\\
			GPV-Pose [CVPR'22]~\cite{di2022gpv}         & \ding{55} & 83.0 & 64.4 & 32.0 & 42.9 & 55.0 & 73.3 &\textbf{8.6}\\
			DPDN [ECCV'22]~\cite{lin2022category}        & \ding{51} & 83.4 & 76.0 & 46.0 & 50.7 & 70.4 & 78.4 &24.6\\
			IST-Net [ICCV'23]~\cite{liu2023net}         & \ding{55} & 82.5 & 76.6 & 47.5 & 53.4 & 72.1 & 80.5 &26.8\\
			Query6DoF [ICCV'23]~\cite{wang2023query6dof}  & \ding{51} & 82.5 & 76.1 & 49.0 & 58.9 & 68.7 & 83.0 &19.7\\
			VI-Net [ICCV'23]~\cite{lin2023vi}            & \ding{55} & -    & -    & 50.0 & 57.6 & 70.8 & 82.1 &28.9\\
			CatFormer [AAAI'24]~\cite{yu2024catformer}    & \ding{51} & 83.1 & 73.8 & 47.7 & 53.7 & 69.0 & 79.5 &-\\
			SecondPose [CVPR'24]~\cite{chen2024secondpose}& \ding{55} & -    & -    & 56.2 & 63.6 & 74.7 & 86.0 &38.1\\  
            AG-Pose [CVPR'24]~\cite{lin2024instance}      & \ding{55} & 83.8 & 77.6 & 56.2 & 62.3 & 73.4 & 81.2 &28.9\\
			AG-Pose (DINOv2)$^\dagger$ [CVPR'24]~\cite{lin2024instance}      & \ding{55} & 84.1 & 80.1 & 57.0 & 64.6 & 75.1 & 84.7 &11.0\\
            GCE-Pose [CVPR'25]~\cite{li2025gce}      & \ding{51} & 84.1 & 79.8 & 57.0 & 65.1 & 75.6 & 86.3 &-\\
            SpherePose [ICLR'25]~\cite{ren2025learning}      & \ding{55} & 84.0 & 79.0 & 58.2 & 67.4 & 76.2 & \textbf{88.2} &-\\
            KeyPose [AAAI'25]~\cite{yu2025keypose}      & \ding{55} & \textbf{84.2} & 80.0 & 57.7 & 66.0 & 78.8 & 88.0 &-\\
            MK-Pose [IROS'25]~\cite{yang2025mk} & \ding{55} & 84.0 & 80.3 & 60.8 & - & 78.0 & 84.6 &180.8\\
            CleanPose [ICCV'25]~\cite{lin2025cleanpose}      & \ding{55} & - & - & 61.7 & 67.6 & 78.3 & 86.3 &14.7\\
			\cline{1-8}
			\textbf{Ours (DINOv2)$^\dagger$}      & \ding{55} & 84.0 & \textbf{80.6} & \textbf{63.2} & \textbf{68.3} & \textbf{80.9} & 88.0 &16.7\\
        \bottomrule
		\end{tabular}}
        \label{tab:real275}
\end{table*}

\renewcommand{\arraystretch}{1.15}
\begin{table*}
    \centering
    \belowrulesep=0pt
    \aboverulesep=0pt
    \setlength{\abovecaptionskip}{0cm}
    \caption{Quantitative comparisons with state-of-the-art methods on the CAMERA25 dataset~\cite{wang2019normalized}. The best results are in \textbf{bold}. The symbol $^\dagger$ indicates that only the trainable parameter count is reported. The symbol $-$ indicates that the data is not available or not reported.}
	\resizebox{0.9\textwidth}{!}{\begin{tabular}{c|c|c c|c c c c c} 
        \toprule
			\multirow{2}{*}{Method} & \multirow{2}{*}{Prior} & \multicolumn{6}{c}{CAMERA25 (\(\uparrow\))} & \#Params \\
			\cline{3-8} 
			& & $\mathrm{IoU_{50}}$ & $\mathrm{IoU_{75}}$ & $5^{\circ}2\mathrm{cm}$ & $5^{\circ}5\mathrm{cm}$ & $10^{\circ}2\mathrm{cm}$ & $10^{\circ}5\mathrm{cm}$ &(M) \(\downarrow\)\\
			\midrule
			SPD [ECCV'20]~\cite{tian2020shape} & \ding{51} & 93.2 & 83.1 & 54.3 & 59.0 & 73.3 & 81.5 &18.3\\
			GPV-Pose [CVPR'22]~\cite{di2022gpv} & \ding{55} & 92.9 & 86.6 & 67.4 & 76.2 & - & 87.4 &\textbf{8.6}\\
			Query6DoF [ICCV'23]~\cite{wang2023query6dof} & \ding{51} & 91.9 & 88.1 & 78.0 & 83.1 & 83.9 & 90.0 &19.7\\
			VI-Net [ICCV'23]~\cite{lin2023vi}&\ding{55}&-&-&74.1&81.4&79.3&87.3 &28.9\\
			CatFormer [AAAI'24]~\cite{yu2024catformer} & \ding{51} & 93.5 & 89.9 & 74.9 & 79.8 & 85.3 & 90.2 &-\\
            AG-Pose [CVPR'24]~\cite{lin2024instance} & \ding{55} & 94.1 & 91.7 & 77.1 & 82.0 & 85.5 & 91.6 &28.9\\
			AG-Pose (DINOv2)$^\dagger$ [CVPR'24]~\cite{lin2024instance} & \ding{55} & 94.2 & 92.5 & 79.5 & 83.7 & 87.1 & 92.6 &11.0\\  
            SpherePose [ICLR'25]~\cite{ren2025learning} & \ding{55} & \textbf{94.8} & 92.4 & 78.3 & 84.3 & 84.8 & 92.3 &-\\
            KeyPose [AAAI'25]~\cite{yu2025keypose}      & \ding{55} & 94.4 & 92.6 & 79.8 & 83.6 & 87.1 & 92.3 &- \\
            MK-Pose [IROS'25]~\cite{yang2025mk} & \ding{55} & 94.1 & 92.2 & 77.9 & - & 86.1 & 91.7 & 180.8 \\
            CleanPose [ICCV'25]~\cite{lin2025cleanpose}      & \ding{55} & - & - & 80.3 & 84.2 & 87.7 & 92.7 &14.7\\
			\cline{1-8}
			\textbf{Ours (DINOv2)$^\dagger$} & \ding{55} & 94.4 & \textbf{92.9} & \textbf{81.0} & \textbf{85.0} & \textbf{88.1} & \textbf{93.4} &16.7\\
            \bottomrule
		\end{tabular}}
        \label{tab:camera25}
\end{table*}

\subsection{Comparison with State-of-the-Art Methods}
\subsubsection{Results on REAL275}

Table~\ref{tab:real275} reports the performance of INKL-Pose compared with state-of-the-art methods on the REAL275 dataset~\cite{wang2019normalized}.
Specifically, INKL-Pose achieves 84.0\% on IoU$_{50}$, 80.6\% on IoU$_{75}$, 63.2\% on 5$^\circ$2cm, 68.3\% on 5$^\circ$5cm, 80.9\% on 10$^\circ$2cm, and 88.0\% on 10$^\circ$5cm, demonstrating its strong capability in both pose and size estimation under real-world conditions.
Compared with GCE-Pose~\cite{li2025gce}, the state-of-the-art prior-based method, INKL-Pose achieves significant improvements of 6.2\% under the strict 5$^\circ$2cm metric.
Furthermore, compared to CleanPose~\cite{lin2025cleanpose}, the current best-performing shape-prior-free method, INKL-Pose yields performance gains of 1.5\% under the strict 5$^\circ$2cm metric, while introducing only a modest increase in model parameters.
In contrast, MK-Pose~\cite{yang2025mk} relies on a substantially larger model, yet does not demonstrate commensurate improvements under strict pose metrics.

\begin{figure*}
	\centering
	\includegraphics[width=0.75\textwidth]{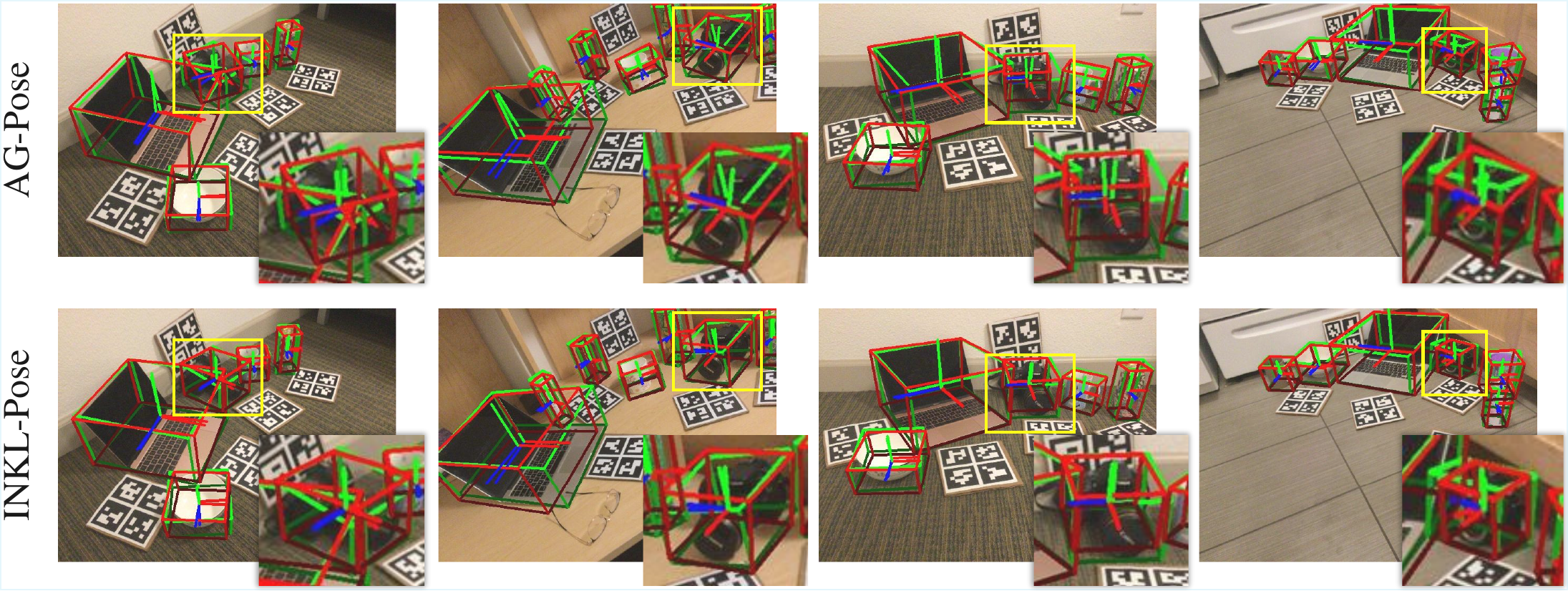}\\
	\caption{Qualitative comparisons between INKL-Pose with AG-Pose~\cite{lin2024instance} on the REAL275 dataset, where green and red represent the predicted and ground-truth bounding boxes, respectively. Our INKL-Pose demonstrates superior precision in pose estimation, even on camera instances with complex geometries.}
	\label{fig:real_vis}
\end{figure*}

In addition, compared to our strong baseline AG-Pose~\cite{lin2024instance}, INKL-Pose achieves notable improvements of 6.2\% on the 5$^\circ$2cm metric, highlighting the effectiveness of our fine-to-global geometric aggregation strategy.
Qualitative comparisons in Fig.~\ref{fig:real_vis} further validate the superiority of INKL-Pose.
Our method accurately estimates object pose even on camera instances with complex geometric structures, while maintaining a favorable balance between accuracy and model complexity.

\subsubsection{Results on CAMERA25}

Table~\ref{tab:camera25} reports the performance of INKL-Pose compared with state-of-the-art methods on the CAMERA25 dataset. As shown in the table, INKL-Pose achieves outstanding accuracy across nearly all evaluation metrics. Although its IoU\(_{50}\) metric reaches 94.4\%, slightly lower than SpherePose~\cite{ren2025learning} (achieving 94.8\%), it significantly outperforms all existing approaches on the remaining metrics, especially under the more rigorous pose accuracy thresholds. Specifically, compared to prior-based methods, INKL-Pose maintains a clear advantage without relying on pre-defined shape priors. In addition, when compared with CleanPose~\cite{lin2025cleanpose}, the current state-of-the-art shape-prior-free method, INKL-Pose achieves notable gains of 0.7\% and 0.8\% under the strict 5°2cm and 5°5cm metrics, respectively. These results, along with the strong performance observed on the REAL275 dataset, provide compelling evidence of the robustness and effectiveness of our proposed local-to-global geometric feature aggregation framework. Furthermore, INKL-Pose exhibits strong model efficiency with 16.7M parameters, achieving a fast inference speed of 36 FPS on an NVIDIA RTX 4090D GPU, highlighting its potential for real-time applications.

\renewcommand{\arraystretch}{1.15}
\begin{table}
\centering
\belowrulesep=0pt
\aboverulesep=0pt
\setlength{\abovecaptionskip}{0cm}
 \caption{Quantitative comparisons with state-of-the-art methods on the HouseCat6D Dataset\cite{jung2024housecat6d}. The best results are in \textbf{bold}.}
 \resizebox{0.5\textwidth}{!}{\begin{tabular}{c|cc|cccc}
 \toprule
	\multirow{2}{*}{Method} & \multicolumn{6}{c}{HouseCat6D (\(\uparrow\))} \\
	\cline{2-7} & IoU$_{25}$ &  IoU$_{50}$ & $5^{\circ}2\mathrm{cm}$ & $5^{\circ}5\mathrm{cm}$ & $10^{\circ}2\mathrm{cm}$ & $10^{\circ}5\mathrm{cm}$\\
	\midrule
	GPV-Pose\cite{di2022gpv}&74.9&50.7&3.5&4.6&17.8&22.7\\
	VI-Net\cite{lin2023vi}&80.7&56.4&8.4&10.3&20.5&29.1\\
	SecondPose\cite{chen2024secondpose}&83.7&66.1&11.0&13.4&25.3&35.7\\
    AG-Pose\cite{lin2024instance}&82.4&66.0&11.5&12.6&37.4&42.5\\
	AG-Pose (DINOv2)\cite{lin2024instance}&88.1&76.9&21.3&22.1&51.3&54.3\\
    SpherePose\cite{ren2025learning}&88.8&72.2&19.3&25.9&40.9&55.3\\
    GCE-Pose\cite{li2025gce}&-&79.2&24.8&25.7&55.4&58.4\\
	\cline{1-7}    \textbf{Ours (DINOv2)}&\textbf{90.8}&\textbf{81.9}&\textbf{25.2}&\textbf{27.4}&\textbf{57.5}&\textbf{61.9}\\ \bottomrule
	\end{tabular}}
    \label{tab:housecat6d_vis}
\end{table}

\subsubsection{Results on HouseCat6D}

\begin{figure}
	\centering
	\includegraphics[width=0.5\textwidth]{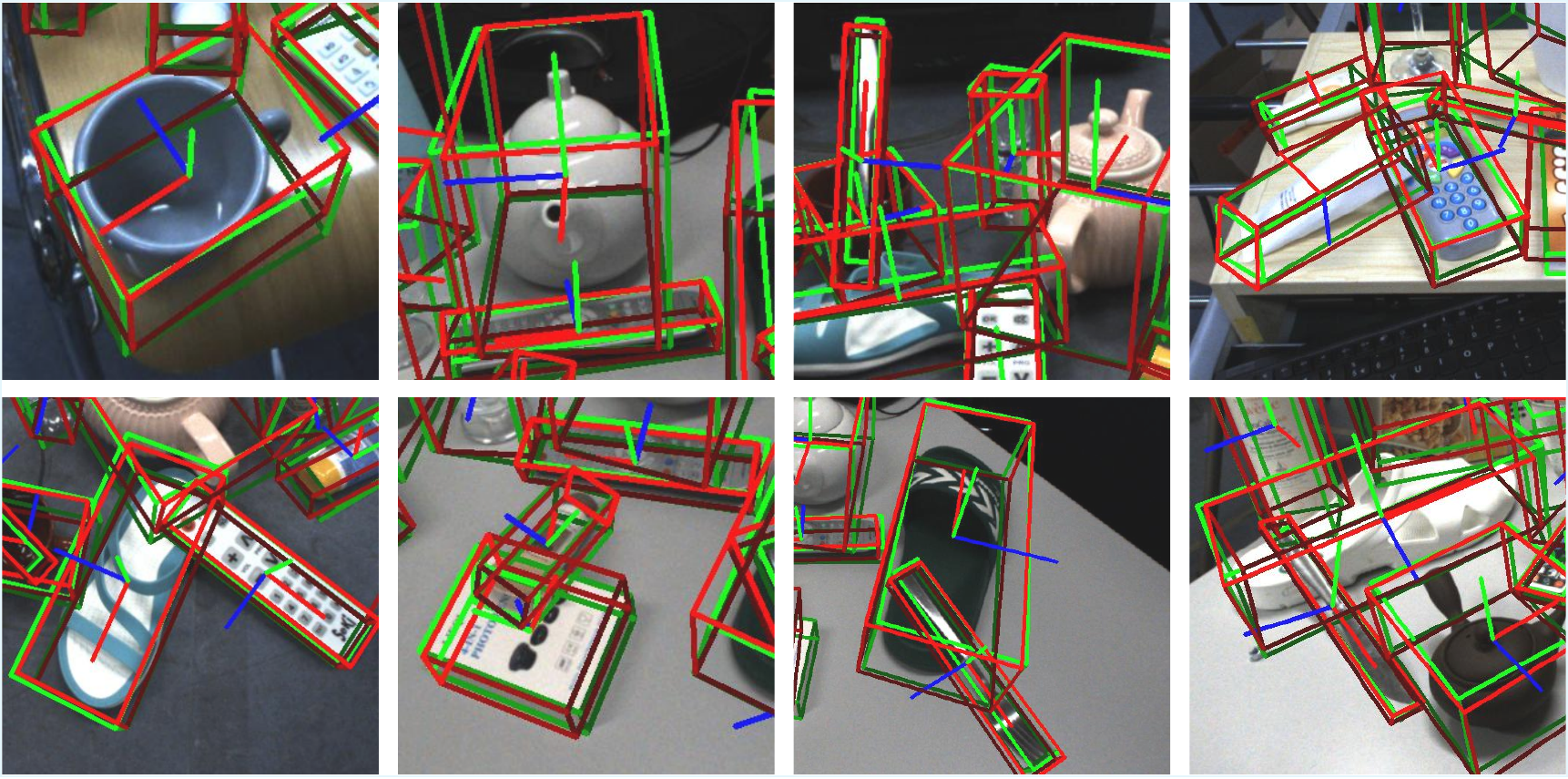}
	\caption{Qualitative visualization results of INKL-Pose on the HouseCat6D~\cite{jung2024housecat6d} dataset. Green bounding boxes represent the ground truth and red bounding boxes represent the predictions of INKL-Pose. The predictions of INKL-Pose remain highly consistent with the ground truth even under occlusion.}
	\label{fig:housecat6d_vis}
\end{figure}

To further demonstrate the effectiveness of INKL-Pose in handling complex real-world scenarios, we conduct experiments on the more challenging real-world dataset HouseCat6D~\cite{jung2024housecat6d}. As reported in Table~\ref{tab:housecat6d_vis}, INKL-Pose achieves state-of-the-art performance across all evaluation metrics. Compared with the current state-of-the-art GCE-Pose~\cite{li2025gce}, our method surpasses it under all evaluation metric, with gains of 2.7\% on IoU$_{50}$, 0.4\% on 5°2cm, 1.7\% on 5°5cm, 2.1\% on 10°2cm, and 3.5\% on 10°5cm. Moreover, Fig.~\ref{fig:housecat6d_vis} shows qualitative pose estimation results under cluttered and heavily occluded conditions, demonstrating that INKL-Pose maintains robust performance in both self-occlusion and inter-object occlusion scenarios.

\begin{figure*}[!htbp]
    \centering
    \includegraphics[width=0.75\textwidth]{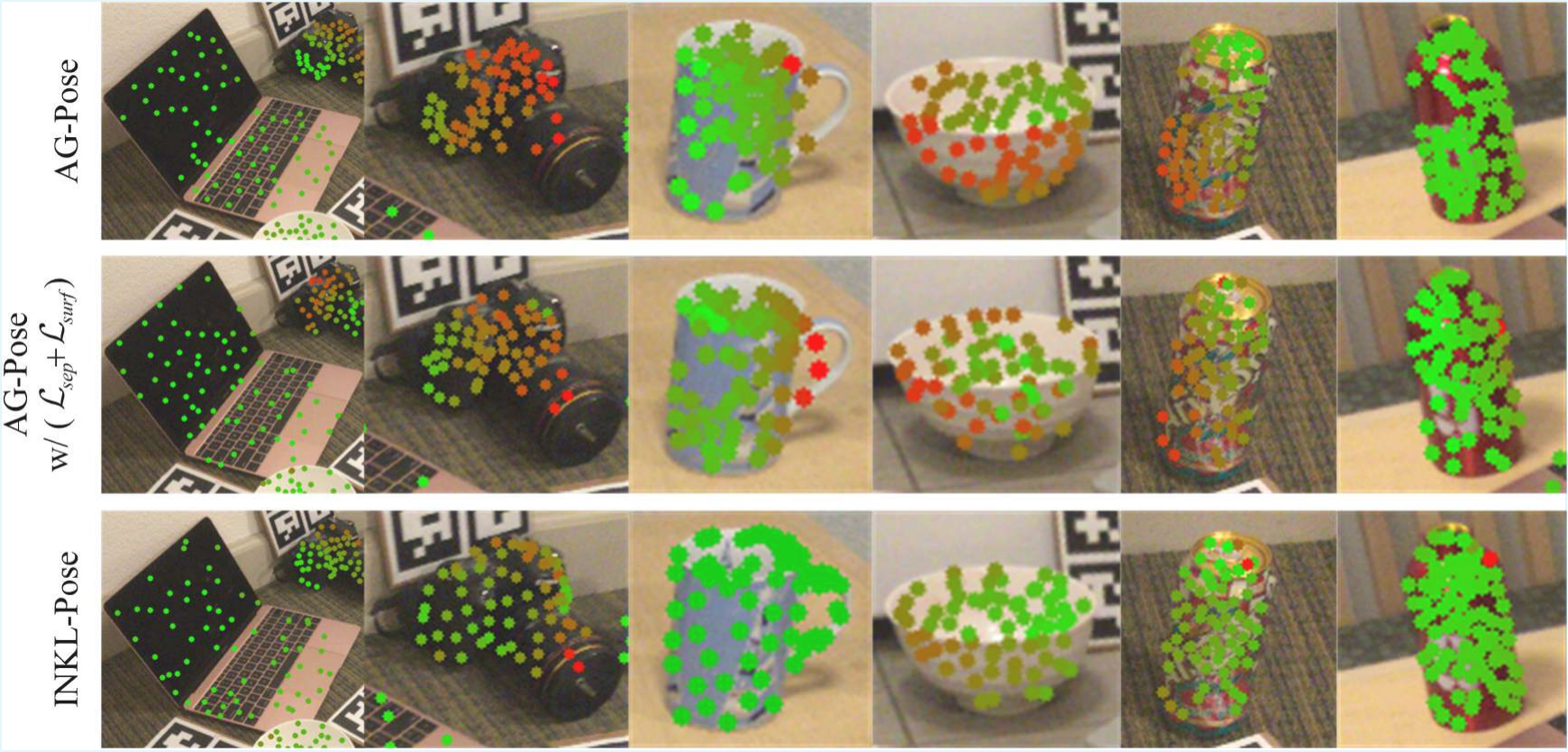}
    \caption{Comparison of predicted keypoints with the baseline method in NOCS space. The first row shows the results of AG-Pose, the second row shows the results of AG-Pose with the proposed keypoint losses, and the third row shows the results of our INKL-Pose. Keypoint colors indicate their NOCS errors, which are linearly mapped within the range \([0, 0.2]\) from green (error = 0) to red (error = 0.2). }
    \label{fig:kpt_cmp}
\end{figure*}

\subsubsection{Visualization of Predicted Keypoints}

Fig.~\ref{fig:kpt_cmp} compares the keypoints predicted by our method with those from the baseline method AG-Pose~\cite{lin2024instance} in the NOCS space. To highlight the differences, keypoints are color-coded based on their NOCS errors, linearly mapped from green (error = 0) to red (error = 0.2). As shown in the figure, AG-Pose produces keypoints with higher NOCS errors and uneven surface coverage. Incorporating our keypoint loss functions into AG-Pose leads to moderate improvements in both accuracy and spatial distribution. In contrast, our INKL-Pose achieves the lowest NOCS errors and generates a more uniform and comprehensive keypoint layout that better captures the object's surface geometry. For example, in the third column showing a mug with a handle, AG-Pose fails to localize the handle, while its variant with our losses partially recovers the region but still exhibits large errors. In comparison, our method accurately delineates the handle contour, resulting in more precise estimations.

\subsection{Ablation Studies}
\label{sec:ablation}

In this section, we conduct comprehensive ablation studies to evaluate the contributions of key components proposed in INKL-Pose on the REAL275~\cite{wang2019normalized} dataset.

\subsubsection{Effect of the Number of Keypoints}
\begin{figure}[H]
    \centering
    \includegraphics[width=0.5\textwidth]{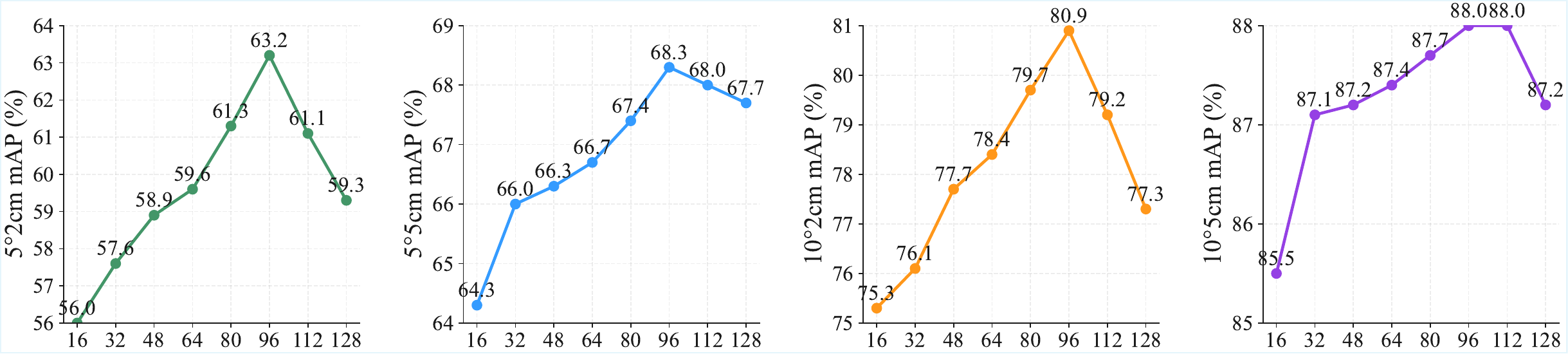}
    \caption{Effect of the number of keypoints.}
    \label{fig:kpt_num}
\end{figure}

We begin by evaluating how the number of predicted keypoints influences the overall pose estimation performance. As shown in Fig.~\ref{fig:kpt_num}, even with only 16 sparse keypoints, INKL-Pose achieves 56.0\% accuracy under the 5$^\circ$2cm metric, which is already comparable to many existing state-of-the-art methods. As the number of keypoints increases, the performance steadily improves due to a stronger capacity for capturing fine-grained geometric details, reaching a peak accuracy of 63.2\% under the strict 5°2cm metric when using 96 keypoints. Beyond this point, further increasing the number of keypoints results in a slight performance decline, caused by redundancy introduced through overly dense keypoint sampling. Therefore, to achieve a balance between computational efficiency and pose estimation accuracy, we adopt 96 keypoints as the default configuration in our framework.



\subsubsection{Effect of the Number of LKFA and GKFA Modules}
\begin{figure}[H]
    \centering
    \includegraphics[width=0.5\textwidth]{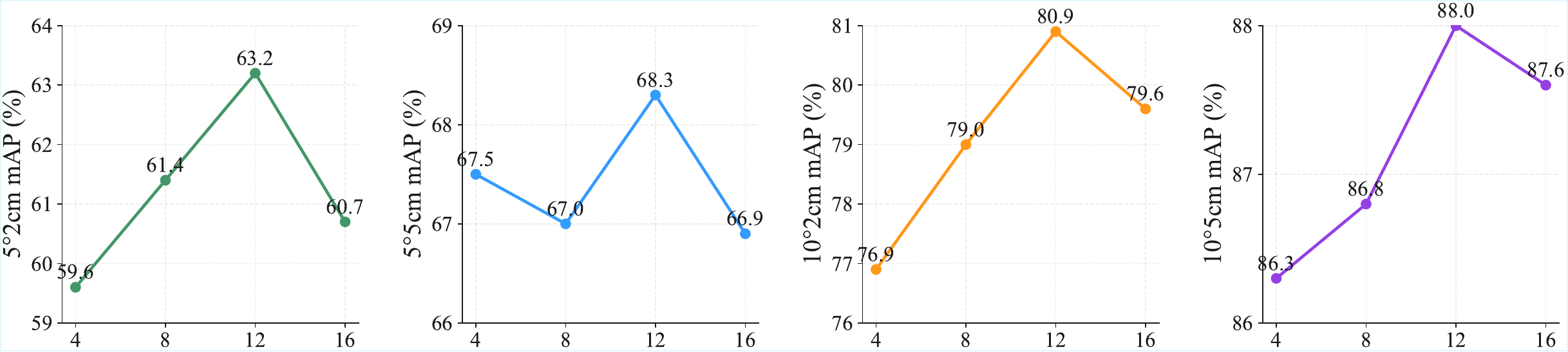}
    \caption{Effect of the number of LKFA and GKFA modules.}
    \label{fig:module_num}
\end{figure}

We investigate the impact of stacking different numbers of keypoint feature aggregation blocks, each containing both a LKFA and a GKFA. As shown in Fig.~\ref{fig:module_num}, increasing the number of blocks from \( S = 4 \) to \( S = 12 \) consistently improves performance across all metrics, as deeper cascaded aggregation enhances both local and global keypoint representations. However, using \( S = 16 \) blocks leads to performance degradation due to feature redundancy, which reduce keypoint distinctiveness and impair pose estimation. Overall, \( S = 12 \) achieves the best balance between accuracy and efficiency.


\subsubsection{Effect of Neighborhood Size}
\begin{figure}[H]
    \centering
    \includegraphics[width=0.5\textwidth]{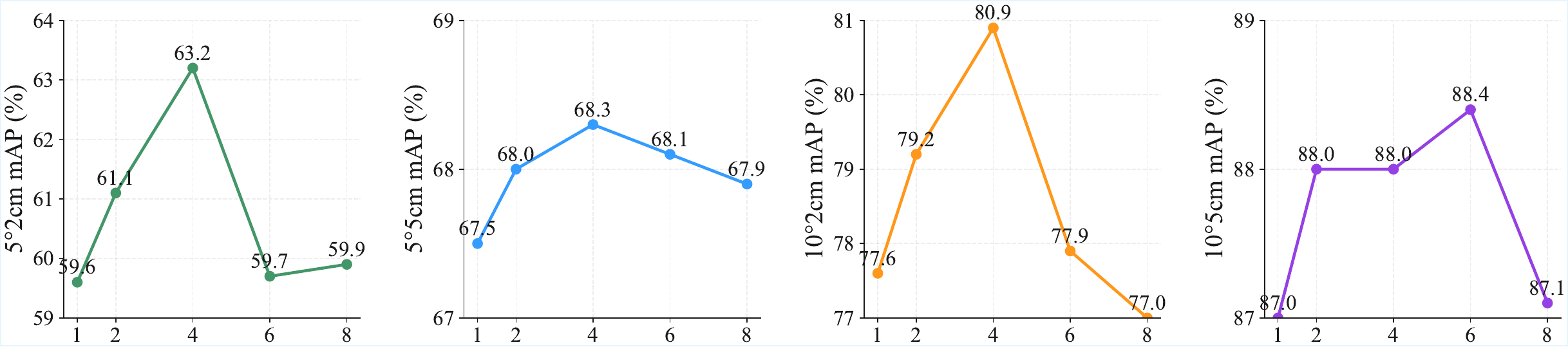}
    \caption{Effect of neighborhood size.}
    \label{fig:k_size}
\end{figure}

To evaluate the influence of neighborhood size \( K \) in the LKFA module, we vary the number of nearest neighbors used during local feature aggregation. As shown in Fig.~\ref{fig:k_size}, when \(K = 1\), the model only uses the center point feature, which limits its ability to capture surrounding geometric and semantic context. Increasing \(K\) enhances local structure modeling, with optimal performance achieved at \(K = 4\). Further increasing \(K\) introduces redundant information, leading to no additional gains.


\subsubsection{Effect of Keypoint Supervision}
In the Instance-Adaptive Keypoint Detector (IAKD) module (Section~\ref{sec:kpt}), we introduce two auxiliary loss terms, \(\mathcal{L}_{surf}\) and \(\mathcal{L}_{sep}\), which encourage uniform surface coverage and spatial diversity among keypoints. Table~\ref{tab:ablation_kpt_loss} summarizes the performance under different supervision strategies. As the table shows, without any keypoint supervision (``None"), the model performs poorly across all metrics, highlighting the necessity of proper guidance. When incorporating the surface coverage loss \(\mathcal{L}_{ocd}\) proposed in AG-Pose, the performance further degrades, indicating that this loss is not well suited to our framework. In contrast, the proposed surface loss \(\mathcal{L}_{surf}\) significantly improves performance over the unsupervised baseline.

For keypoint separation, both the \(\mathcal{L}_{div}\) loss from AG-Pose and our proposed \(\mathcal{L}_{sep}\) result in clear performance gains, although \(\mathcal{L}_{sep}\) consistently achiev\-es higher accuracy. Notably, the combination of \(\mathcal{L}_{surf}\) and \(\mathcal{L}_{sep}\) delivers the best performance across all metrics, demonstrating their complementary contributions to keypoint learning. These results confirm that the proposed losses more effectively guide the network to generate semantically consistent, geometrically meaningful, and spatially well-distributed keypoints, which ultimately enhances pose estimation performance.

\subsubsection{Effect of Positional Embeddings}
We further analyze the contribution of positional information to feature representation. In INKL-Pose, positional encodings are introduced at two distinct stages: the embedding \(\mathcal{E}_{\mathcal{P}}\) in the Pointwise Feature Encoder (Section~\ref{sec:fea_extractor}), which compensates for the loss of absolute position cues, and \(\mathcal{E}_{kpt}\) in the LKFA module (Section~\ref{sec:LKFA}), which enhances local spatial awareness. As reported in Table~\ref{tab:pos_emb}, removing either embedding individually leads to a noticeable performance drop across all metrics, highlighting their complementary roles in capturing geometric information.

\begin{table}[ht]
	\centering
    \belowrulesep=0pt
    \aboverulesep=0pt
    \setlength{\abovecaptionskip}{0cm}
    \caption{Effect of keypoint supervision. Default configurations are colored in gray, with best results highlighted in \textbf{bold}.}
	\begin{tabular}{c|cccc} 
		\toprule
		Method & 5$^\circ$2cm & 5$^\circ$5cm & 10$^\circ$2cm & 10$^\circ$5cm \\
		\midrule  \rowcolor{gray!30}
		w/ (\(\mathcal{L}_{sep}\)+\(\mathcal{L}_{surf}\)) & \textbf{63.2} & \textbf{68.3} & \textbf{80.9} & \textbf{88.0} \\
		w/ (\(\mathcal{L}_{div}\)+\(\mathcal{L}_{ocd}\)) & 60.9 & 67.0 & 79.9 & 87.2 \\ 
		\midrule
		w/ \(\mathcal{L}_{sep}\) (Ours) & 60.8 & 67.7 & 77.8 & 86.2 \\
		w/ \(\mathcal{L}_{div}\) (AG-Pose~\cite{lin2024instance}) &59.1 &66.9 &77.2 &86.0\\
		\midrule
		w/ \(\mathcal{L}_{surf}\) (Ours) & 46.9 & 53.0 & 69.1 & 78.8 \\
		w/ \(\mathcal{L}_{ocd}\) (AG-Pose~\cite{lin2024instance}) &11.8 &18.3 &24.7 &45.6\\
		\midrule
		None &18.5 &20.4 &43.7 &53.3\\
		\bottomrule
	\end{tabular}
	\label{tab:ablation_kpt_loss}
\end{table}

\begin{table}[ht]
\centering
\belowrulesep=0pt
\aboverulesep=0pt
\setlength{\abovecaptionskip}{0cm}
 \caption{Effect of positional embeddings. Default configurations are colored in gray, with best results highlighted in \textbf{bold}.}
 \begin{tabular}{c|cccc}  
    \toprule
	Method & 5$^\circ$2cm & 5$^\circ$5cm & 10$^\circ$2cm & 10$^\circ$5cm \\
	\hline  \rowcolor{gray!30}
	Full & \textbf{63.2} & \textbf{68.3} & \textbf{80.9} & \textbf{88.0} \\
	\midrule
	w/o \(\mathcal{E}_{\mathcal{P}}\) & 60.4 & 67.1 & 77.6 & 86.2 \\
	w/o $\mathcal{E}_{kpt}$ & 59.1 & 66.6 & 76.9 & 85.8 \\
    \bottomrule
	\end{tabular}
    \label{tab:pos_emb}
\end{table}

\subsubsection{Effect of the LKFA and GKFA Modules}
We evaluate the contributions of the LKFA and GKFA modules introduced in Section~\ref{sec:LKFA} and Section~\ref{sec:GKFA}, respectively. As shown in Table~\ref{tab:modules}, removing the LKFA results in a 2.1\% drop in the stringent 5°2cm metric, primarily due to the loss of fine-grained geometric features. Furthermore, removing the GKFA leads to a larger decline of 3.0\%, indicating its role in preserving global consistency. These results underscore that both local and global geometric information are crucial for establishing accurate correspondences on unseen objects within the same category.

\begin{table}[ht]
\centering
\belowrulesep=0pt
\aboverulesep=0pt
\setlength{\abovecaptionskip}{0cm}
 \caption{Effect of the LKFA and GKFA modules. Default configurations are colored in gray, with best results highlighted in \textbf{bold}.}
 \begin{tabular}{c|cccc} 
    \toprule
	Method & 5$^\circ$2cm & 5$^\circ$5cm & 10$^\circ$2cm & 10$^\circ$5cm \\
	\midrule  \rowcolor{gray!30}
	Full & \textbf{63.2} & \textbf{68.3} & \textbf{80.9} & 88.0 \\
	\midrule
	w/o LKFA & 61.1 & 67.2 & 80.5 & \textbf{88.4} \\
	w/o GKFA & 60.2 & 67.6 & 77.1 & 86.6 \\
    \bottomrule
	\end{tabular}
    \label{tab:modules}
\end{table}

\subsubsection{Effect of Bidirectional Mamba and Feature Sequence Flipping}
We also examine the contributions of the bidirectional Mamba architecture and the proposed Feature Sequence Flipping (FSF) strategy within the GKFA module (Section~\ref{sec:GKFA}), as summarized in Table~\ref{tab:gkfa}. The results show that replacing the bidirectional Mamba with its unidirectional counterpart (Uni-Mamba) consistently degrades performance, confirming the advantage of bidirectional modeling for global feature aggregation. Furthermore, we replace the bidirectional Mamba in GKFA with a standard attention-based module consisting of a self-attention layer followed by a feed-forward network (FFN), corresponding to the core sequence modeling component in Transformer architectures. Under identical architectural and training settings, this attention-based variant exhibits a clear performance drop, particularly under the strict $5^\circ$2cm metric, and does not outperform the unidirectional Mamba baseline, indicating that Mamba is more effective for capturing long-range dependencies among keypoints in this setting. Additionally, we compare FSF with the commonly used Point Sequence Flipping (PSF) strategy. PSF exacerbates the non-causal nature of point clouds by reversing the sequence along spatial indices, disrupting the alignment between feature order and geometric structure, and leading to performance drops even below the unidirectional baseline. In contrast, FSF preserves the original point order while achieving bidirectional information flow in feature space, effectively improving robustness to the unordered and non-causal characteristics of point clouds.

\begin{table}[ht]
\centering
\belowrulesep=0pt
\aboverulesep=0pt
\setlength{\abovecaptionskip}{0cm}
\caption{Effect of bidirectional Mamba and feature sequence flipping. Default configurations are colored in gray, with best results highlighted in \textbf{bold}.}
\begin{tabular}{c|cccc} 
 \toprule
 Method & 5$^\circ$2cm & 5$^\circ$5cm & 10$^\circ$2cm & 10$^\circ$5cm \\
	\midrule  \rowcolor{gray!30}
    w/ Bi-Mamba   & \textbf{63.2} & \textbf{68.3} & \textbf{80.9} & \textbf{88.0} \\
	w/ Uni-Mamba   & 60.0 & 66.6 & 78.5 & 86.8 \\
    w/ Attention   & 59.8 & 66.8 & 78.8 & 86.8 \\
    \midrule
	PSF & 58.2 & 65.2 & 77.0 & 85.2 \\  \rowcolor{gray!30}
	FSF & \textbf{63.2} & \textbf{68.3} & \textbf{80.9} & \textbf{88.0} \\
 \bottomrule
\end{tabular}
\label{tab:gkfa}
\end{table}

\begin{figure}[ht]
	\centering    
    \includegraphics[width=0.5\textwidth]{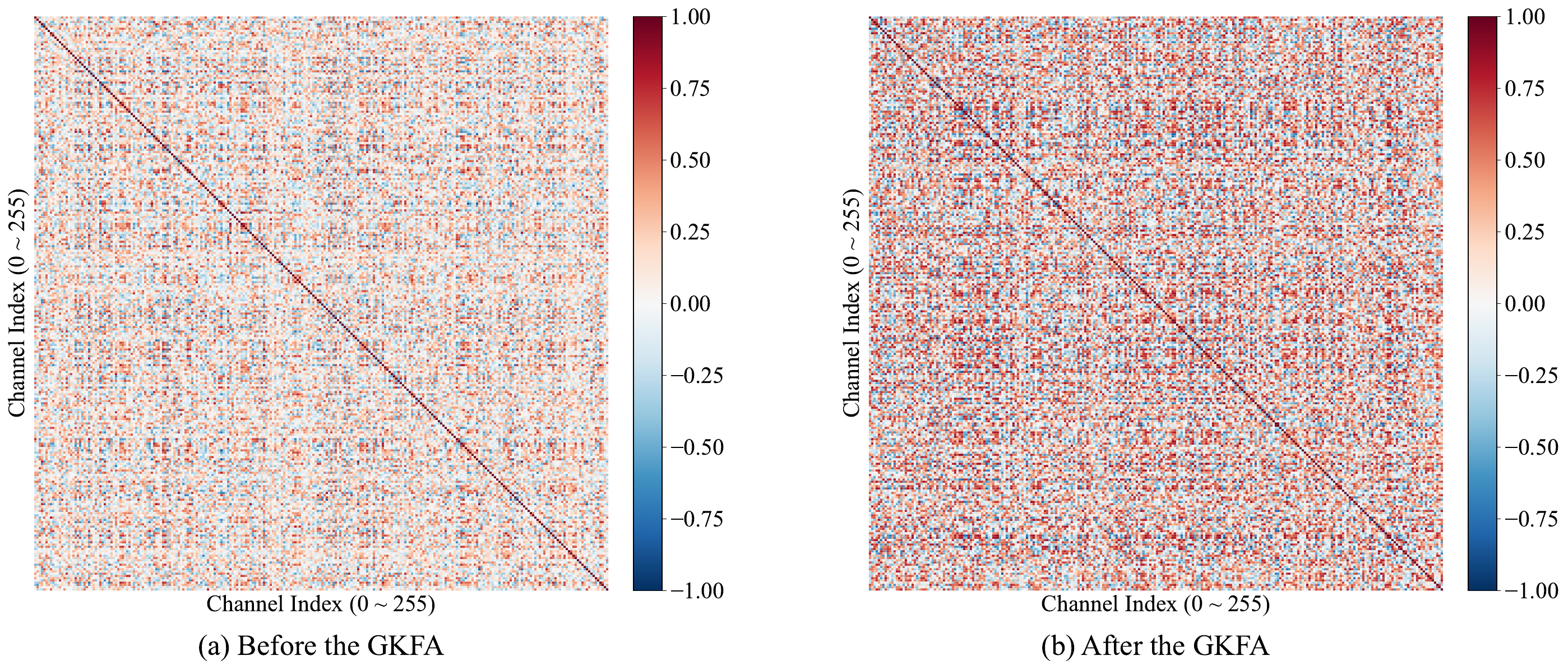}
	\caption{Visualization of cosine similarity matrices of keypoint feature channels. (a) Channel-wise similarity of local keypoint features before GKFA. (b) Channel-wise similarity after GKFA.}
	\label{fig:fsf}
\end{figure}

To further analyze the behavior of FSF within the GKFA module, we provide a qualitative visualization of channel-wise feature correlations in Fig.~\ref{fig:fsf}.
Specifically, Fig.~\ref{fig:fsf}(a) shows the cosine similarity matrix of keypoint feature channels before global aggregation, where structured but uneven inter-channel relationships can be observed, reflecting the heterogeneous nature of local geometric and semantic cues.
After applying GKFA with bidirectional Mamba and FSF, Fig.~\ref{fig:fsf}(b) exhibits a more coherent and structured correlation pattern across feature channels.
This suggests that bidirectional aggregation in feature space facilitates more effective information interaction among channels, which is consistent with the quantitative improvements reported in Table~\ref{tab:gkfa}.

\section{Conclusion}
\label{sec:conclusion}

In this paper, we propose INKL-Pose, a novel keypoint-based framework for category-level object pose estimation that enhances generalization to unseen object instances within the same category through instance-adaptive keypoint learning and local-to-global geometric aggregation. Specifically, we introduce an Instance-Adaptive Keypoint Detector that predicts semantically consistent and geometrically informative keypoints, along with two complementary modules for feature refinement: a Local Keypoint Feature Aggregator that captures fine-grained geometric details and a Global Keypoint Feature Aggregator based on bidirectional Mamba with a Feature Sequence Flipping strategy for structure-aware global perception. Furthermore, we design a surface loss and a separation loss to explicitly supervise keypoint distribution, providing strong geometric guidance for robust pose estimation. Extensive experiments on CAMERA25, REAL275, and HouseCat6D demonstrate that INKL-Pose achieves state-of-the-art performance across both synthetic and real-world datasets, significantly outperforming existing approaches.
Despite these promising results, our method still faces several limitations.
First, the current framework operates on static single-frame observations and does not explicitly exploit temporal consistency across consecutive frames, which may limit its robustness in dynamic or continuous motion scenarios.
Second, as also reflected in the qualitative failure cases, performance can be affected by challenging optical conditions such as high reflectivity or transparency, where reliable semantic feature extraction becomes difficult.
Third, severe occlusion and atypical object geometries that deviate significantly from common category-level shape patterns may lead to ambiguous or incomplete keypoint representations.
Finally, since INKL-Pose relies on accurate keypoint detection to encode object structure, errors in keypoint localization can propagate to subsequent pose estimation.
In future work, we plan to extend INKL-Pose to temporal settings for sequential pose tracking, and to explore more robust semantic feature extraction and uncertainty-aware keypoint modeling, aiming to improve performance under challenging optical conditions, severe occlusion, and complex object geometries.

\bibliographystyle{IEEEtran}
\bibliography{ref}

\vfill

\clearpage
\begin{bibunit}[IEEEtran]

\section*{Appendix}
\renewcommand{\thesection}{\Roman{section}}
\setcounter{section}{0}

\setcounter{table}{0}
\renewcommand{\thetable}{\Roman{table}}
\renewcommand{\theHtable}{appendix.\Roman{table}}

\setcounter{figure}{0}
\renewcommand{\thefigure}{\arabic{figure}}
\renewcommand{\theHfigure}{appendix.\arabic{figure}}

\section{Sensitivity Analysis of Loss Weights}
In our training configuration, we design two novel loss functions: the separation loss $\mathcal{L}_{sep}$, which encourages uniform distribution of keypoints, and the surface loss $\mathcal{L}_{surf}$, which constrains keypoints to lie on the object surface. Their weights are set to $w_{sep} = 10.0$ and $w_{surf} = 10.0$, respectively. For the remaining three loss functions, namely the similarity loss $\mathcal{L}_{sim}$, the mapping loss $\mathcal{L}_{map}$, and the pose loss $\mathcal{L}_{pose}$, we follow AG-Pose~\cite{lin2024instance} and set the weights to $w_{sim} = 15.0$, $w_{map} = 2.0$, and $w_{pose} = 0.3$. These values are derived from empirical observations in AG-Pose, and our experiments demonstrate that they are also applicable to our method. To further investigate the influence of loss weights on model performance, we report experimental results with varying values of $w_{sep}$ and $w_{surf}$.

Regarding $w_{sep}$, we conduct experiments on REAL275~\cite{wang2019normalized} with values of 1.0, 2.0, 4.0, 8.0, and 15.0, while keeping all other loss terms and training settings unchanged. As shown in Table~\ref{tab:Lsep_real}, $w_{sep}=10.0$ appears to be the optimal choice for most evaluation metrics, with relatively minor performance fluctuations across different settings.

\begin{table}[ht]
\centering
\belowrulesep=0pt
\aboverulesep=0pt
\setlength{\abovecaptionskip}{0cm}
\caption{Sensitivity analysis of the weight for the separation loss $\mathcal{L}_{sep}$ on REAL275~\cite{wang2019normalized}. The best results are in \textbf{bold}.}
\resizebox{0.5\textwidth}{!}{\begin{tabular}{c|c c c c}
\hline
\multirow{2}{*}{$w_{sep}:$ $w_{surf}:$ $w_{sim}:$ $w_{map}:$ $w_{pose}$} &
\multicolumn{4}{c}{REAL275 (\(\uparrow\))} \\
\cline{2-5}
 & 5$^\circ$2cm & 5$^\circ$5cm & 10$^\circ$2cm & 10$^\circ$5cm \\
\hline
1.0 : 10.0 : 15.0 : 2.0 : 0.3  & 59.3 & 66.2 & 76.9 & 85.3 \\
2.0 : 10.0 : 15.0 : 2.0 : 0.3  & 59.7 & 67.2 & 77.9 & 86.6 \\
4.0 : 10.0 : 15.0 : 2.0 : 0.3  & 60.2 & 67.0 & 78.5 & 87.6 \\
8.0 : 10.0 : 15.0 : 2.0 : 0.3  & 62.5 & 68.2 & 80.7 & 87.5 \\
\textbf{10.0 : 10.0 : 15.0 : 2.0 : 0.3} & \textbf{63.2} & \textbf{68.3} & \textbf{80.9} & \textbf{88.0} \\
15.0 : 10.0 : 15.0 : 2.0 : 0.3 & 61.4 & 67.0 & 79.0 & 86.8 \\
\hline
\end{tabular}}
\label{tab:Lsep_real}
\end{table}

Regarding $w_{surf}$, we conduct experiments on REAL275~\cite{wang2019normalized} with values of 1.0, 2.0, 4.0, 8.0, and 15.0, while keeping all other loss terms and training settings unchanged. As shown in Table~\ref{tab:Lsurf_real}, $w_{surf}=10.0$ appears to be the optimal choice for most evaluation metrics, with relatively minor performance fluctuations across different settings.

\renewcommand{\arraystretch}{1.15}
\begin{table}[ht]
\centering
\belowrulesep=0pt
\aboverulesep=0pt
\setlength{\abovecaptionskip}{0cm}
\caption{Sensitivity analysis of the weight for the surface loss $\mathcal{L}_{surf}$ on REAL275~\cite{wang2019normalized}. The best results are in \textbf{bold}.}
\resizebox{0.5\textwidth}{!}{\begin{tabular}{c|c c c c}
\hline
\multirow{2}{*}{$w_{sep}:$ $w_{surf}:$ $w_{sim}:$ $w_{map}:$ $w_{pose}$} &
\multicolumn{4}{c}{REAL275 (\(\uparrow\))} \\
\cline{2-5}
 & 5$^\circ$2cm & 5$^\circ$5cm & 10$^\circ$2cm & 10$^\circ$5cm \\
\hline
10.0 : 1.0 : 15.0 : 2.0 : 0.3  & 60.7 & 67.4 & 77.6 & 87.3 \\
10.0 : 2.0 : 15.0 : 2.0 : 0.3  & 61.3 & 67.0 & 79.0 & 87.3 \\
10.0 : 4.0 : 15.0 : 2.0 : 0.3  & 61.7 & 66.5 & 80.4 & 87.2 \\
10.0 : 8.0 : 15.0 : 2.0 : 0.3  & 61.9 & \textbf{69.1} & 78.4 & 87.2 \\
\textbf{10.0 : 10.0 : 15.0 : 2.0 : 0.3} & \textbf{63.2} & 68.3 & \textbf{80.9} & \textbf{88.0} \\
10.0 : 15.0 : 15.0 : 2.0 : 0.3 & 61.3 & 67.4 & 79.7 & 87.7 \\
\hline
\end{tabular}}
\label{tab:Lsurf_real}
\end{table}

\section{Computational Efficiency and Memory Usage Analysis}
To evaluate computational efficiency and memory usage, we conduct a comparative analysis among AG-Pose~\cite{lin2024instance}, INKL-Pose with bidirectional Mamba (our full model), and an attention-based variant of INKL-Pose.
In the variant model, the bidirectional Mamba in the GKFA is replaced with a standard attention module, while all LKFA modules and other components remain identical.
The evaluation metrics include training time per epoch, total inference time, FLOPs, GPU memory consumption, and parameter count.
For inference speed, we perform 10 runs on the REAL275 test set consisting of 2,754 frames and report the averaged total inference time.
All experiments are conducted with a batch size of 40 for fair comparison.

The results are summarized in Table~\ref{tab:efficiency}.
All three methods exhibit comparable training overhead, requiring approximately 18.4 minutes per epoch.
Although INKL-Pose is slightly slower than AG-Pose during inference due to the introduced multi-level local-to-global geometric aggregation, it achieves significantly improved pose accuracy.
Notably, INKL-Pose with Mamba shows the lowest FLOPs and GPU memory consumption among the three models.
Furthermore, compared with the attention-based variant, the Mamba-based INKL-Pose achieves faster inference, lower model complexity, reduced memory usage, and fewer parameters, while also delivering superior accuracy.
These results further support the efficiency advantages of adopting Mamba for modeling long-range dependencies among keypoints.

\renewcommand{\arraystretch}{1.15}
\begin{table}[ht]
\centering
\belowrulesep=0pt
\aboverulesep=0pt
\setlength{\abovecaptionskip}{0cm}
\caption{Comparison of training cost, inference time, FLOPs, GPU memory usage, and parameter counts across different methods.}
\resizebox{0.5\textwidth}{!}{\begin{tabular}{c|c|cc}
\hline
 & \multirow{2}{*}{AG-Pose~\cite{lin2024instance}}
 & \multicolumn{2}{c}{INKL-Pose} \\
\cline{3-4}
 &  & w/ Mamba & w/ Attention \\
\hline
Training time (min / epoch) & 18.4 & 18.4 & 18.4 \\
Total inference time (s)          & 61.2 & 76.5 & 83.4 \\
FLOPs (GFLOPs) & 60.2 & 47.5 & 52.1 \\
Memory usage (MiB)          & 12184 & 11594 & 11726 \\
Parameters (M)              & 11.0 & 16.7 & 20.4 \\
\hline
\end{tabular}}
\label{tab:efficiency}
\end{table}

\section{Per-Category Comparison Results on REAL275}

To provide a more comprehensive and fine-grained evaluation, we report the category-wise performance of INKL-Pose on the REAL275~\cite{wang2019normalized} dataset in Table~\ref{tab:category-wise_real}.
This comparison includes both symmetric and asymmetric object categories and highlights the robustness of our method under varying levels of pose ambiguity.
Overall, INKL-Pose consistently outperforms the baseline AG-Pose~\cite{lin2024instance} across most categories and evaluation metrics.

For symmetric objects, INKL-Pose demonstrates clear advantages on several challenging categories.
Specifically, for the Bottle category, INKL-Pose improves the strict $5^{\circ}$2cm metric by 6.0\%.
For the Mug category, which exhibits severe pose ambiguity due to handle orientation, INKL-Pose achieves a substantial improvement of 15.4\% under the strict $5^{\circ}$2cm criterion.
For the Bowl and Can categories, both methods achieve comparable performance, indicating that INKL-Pose does not sacrifice accuracy on relatively simpler symmetric cases.
For asymmetric objects, INKL-Pose exhibits even more pronounced performance gains.
In the Camera category, which is particularly challenging due to complex geometric structures, INKL-Pose substantially outperforms AG-Pose, improving the $5^{\circ}$2cm, $10^{\circ}$2cm, and $10^{\circ}$5cm metrics from 1.4\% to 6.3\%, from 28.7\% to 44.2\%, and from 34.1\% to 49.3\%, respectively.
Similarly, for the Laptop category, INKL-Pose achieves a notable improvement of 10.9\% under the strict $5^{\circ}$2cm metric.

On average, INKL-Pose outperforms AG-Pose by 6.2\%, 3.7\%, 5.8\%, and 3.3\% under the $5^{\circ}$2cm, $5^{\circ}$5cm, $10^{\circ}$2cm, and $10^{\circ}$5cm metrics, respectively.
These consistent improvements across evaluation criteria and object categories demonstrate the robustness of the proposed method, particularly in handling ambiguous symmetric objects.

\begin{table*}[ht]
    \centering
    \belowrulesep=0pt
    \aboverulesep=0pt
    \setlength{\abovecaptionskip}{0cm}
    \caption{Overall and category-wise evaluation on REAL275~\cite{wang2019normalized}. The best results are in \textbf{bold}.}
    \begin{tabular}{c c c c c c c c c}
    \toprule
    \multirow{1}{*}{} & Categories & Method & IoU$_{50}$ & IoU$_{75}$ & 5$^\circ$2cm & 5$^\circ$5cm & 10$^\circ$2cm & 10$^\circ$5cm \\
    \midrule

    \multirow{8}{*}{Symmetry}
    & \multirow{2}{*}{Bottle}
    & AG-Pose~\cite{lin2024instance} & 57.5 & 50.1 & 72.6 & 79.8 & 78.9 & 87.6 \\
    & & Ours & \textbf{57.7} & \textbf{50.1} & \textbf{78.6} & \textbf{82.3} & \textbf{82.2} & \textbf{87.8} \\
    \cmidrule(lr){2-9}

    & \multirow{2}{*}{Bowl}
    & AG-Pose~\cite{lin2024instance} & 100 & 100 & \textbf{93.8} & \textbf{98.9} & 94.7 & 99.9 \\
    & & Ours & \textbf{100} & \textbf{100} & 93.7 & 98.0 & \textbf{95.8} & \textbf{100} \\
    \cmidrule(lr){2-9}

    & \multirow{2}{*}{Can}
    & AG-Pose~\cite{lin2024instance} & 71.4 & 71.2 & \textbf{79.8} & \textbf{82.6} & 95.9 & 98.5 \\
    & & Ours & \textbf{71.4} & \textbf{71.3} & 79.7 & 82.1 & \textbf{95.9} & \textbf{98.7} \\
    \cmidrule(lr){2-9}

    & \multirow{2}{*}{Mug}
    & AG-Pose~\cite{lin2024instance} & \textbf{99.6} & 99.2 & 34.6 & 34.8 & 89.4 & 89.9 \\
    & & Ours & 99.5 & \textbf{99.2} & \textbf{50.0} & \textbf{50.2} & \textbf{94.8} & \textbf{95.1} \\
    \cdashline{1-9}

    \multirow{4}{*}{Asymmetry}
    & \multirow{2}{*}{Camera}
    & AG-Pose~\cite{lin2024instance} & 90.9 & 84.8 & 1.4 & 1.7 & 28.7 & 34.1 \\
    & & Ours & \textbf{90.9} & \textbf{86.4} & \textbf{6.3} & \textbf{6.7} & \textbf{44.2} & \textbf{49.3} \\
    \cmidrule(lr){2-9}

    & \multirow{2}{*}{Laptop}
    & AG-Pose~\cite{lin2024instance} & \textbf{85.2} & 76.4 & 59.8 & 89.7 & 63.1 & \textbf{97.6} \\
    & & Ours & 84.6 & \textbf{76.5} & \textbf{70.7} & \textbf{90.6} & \textbf{72.5} & 96.9 \\
    \cmidrule(lr){1-9}

    & \multirow{2}{*}{Average}
    & AG-Pose~\cite{lin2024instance} & 84.1 & 80.1 & 57.0 & 64.6 & 75.1 & 84.7 \\
    & & Ours & \textbf{84.1} & \textbf{80.6} & \textbf{63.2} & \textbf{68.3} & \textbf{80.9} & \textbf{88.0} \\
    \bottomrule
    \end{tabular}
    \label{tab:category-wise_real}
\end{table*}

\section{Qualitative Failure Cases and Limitations}

\begin{figure}[ht]
    \centering
    \includegraphics[width=\linewidth]{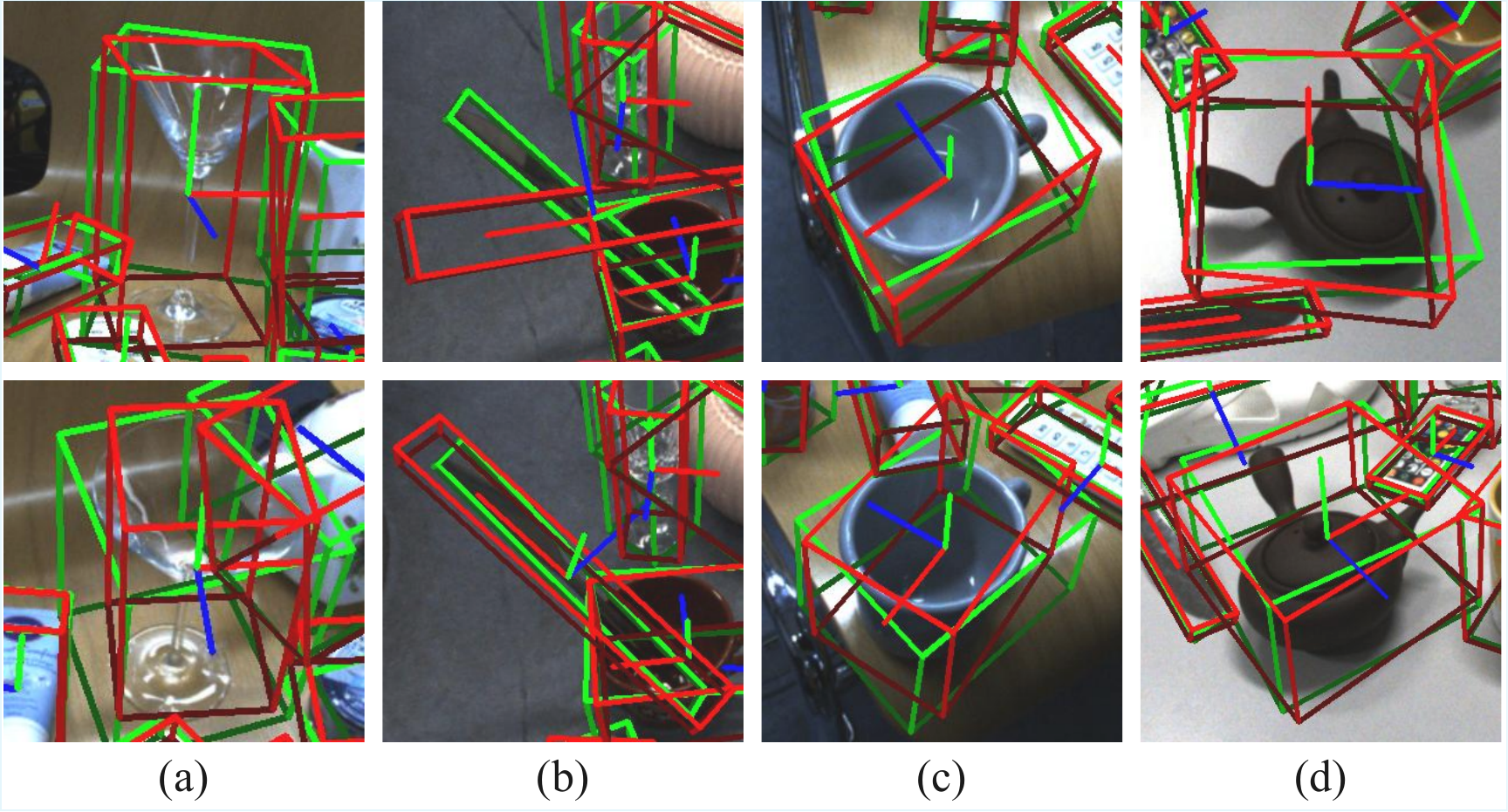}
    \caption{Failure cases on the HouseCat6D~\cite{jung2024housecat6d} dataset: (a) transparent objects; (b) highly reflective tubular structures; (c) severe self-occlusion; (d) atypical object geometries.}
    \label{fig:failcase}
\end{figure}

Fig.~\ref{fig:failcase} presents representative failure cases on the HouseCat6D~\cite{jung2024housecat6d} dataset, which can be broadly grouped into several typical scenarios.
Specifically, (a) shows transparent objects, where DINOv2~\cite{oquab2023dinov2} fails to extract reliable semantic features, leading to degraded pose estimation;
(b) illustrates highly reflective tubular objects that similarly hinder robust feature extraction;
(c) corresponds to severe self-occlusion, where key structural cues (e.g., handles) are largely invisible, increasing pose ambiguity; and
(d) depicts failures caused by atypical object geometries that deviate from common category-level shape patterns.
Overall, these limitations highlight open challenges for category-level pose estimation and point to potential directions for future improvement.

\putbib[ref]
\end{bibunit}

\end{document}